\pdfoutput=1

\PassOptionsToPackage{table}{xcolor}
\documentclass[11pt]{article}

\usepackage[final]{acl}

\usepackage{times}
\usepackage{latexsym}

\usepackage[T1]{fontenc}

\usepackage[utf8]{inputenc}

\usepackage{microtype}

\usepackage{inconsolata}

\usepackage{graphicx}
\usepackage{amsmath}
\usepackage{booktabs}
\usepackage{multirow}
\usepackage{tabularx}
\usepackage{amssymb}
\usepackage{pifont}
\newcommand{\cmark}{\ding{51}}
\newcommand{\xmark}{\ding{55}}
\usepackage{enumitem}
\usepackage{xcolor}
\usepackage[breakable,skins]{tcolorbox}
\usepackage{mdframed}
\usepackage{hyperref}
\usepackage{subcaption}
\hypersetup{pdftitle={From What to Respond to When to Respond: Timely Response Generation for Open-domain Dialogue Agents},
            pdfauthor={Seongbo Jang, Minjin Jeon, Jaehoon Lee, Seonghyeon Lee, Dongha Lee, Hwanjo Yu}}

\title{From \textit{What to Respond} to \textit{When to Respond}: \\Timely Response Generation for Open-domain Dialogue Agents}

\author{
Seongbo Jang\textsuperscript{1} \quad
Minjin Jeon\textsuperscript{2} \quad
Jaehoon Lee\textsuperscript{3} \\
\textbf{Seonghyeon Lee\textsuperscript{4} \quad
Dongha Lee\textsuperscript{5,*} \quad
Hwanjo Yu\textsuperscript{2,*}} \\
\textsuperscript{1}Myongji University \quad
\textsuperscript{2}Pohang University of Science and Technology \\
\textsuperscript{3}Scatter Lab, Inc. \quad
\textsuperscript{4}Kyungpook National University \quad
\textsuperscript{5}Yonsei University \\
\texttt{sjang@mju.ac.kr} \quad
\texttt{\{minjinj,hwanjoyu\}@postech.ac.kr} \\
\texttt{hyudlwogns0128@gmail.com} \quad
\texttt{sh0416@knu.ac.kr} \quad
\texttt{donalee@yonsei.ac.kr}
}

\begin{document}
\maketitle
{\renewcommand{\thefootnote}{\fnsymbol{footnote}}\footnotetext[1]{Corresponding authors}}

\begin{abstract}
While research on dialogue response generation has primarily focused on generating coherent responses conditioning on textual context, the critical question of \textit{when to respond} grounded on the temporal context remains underexplored.
To bridge this gap, we propose the task of timely dialogue response generation and introduce the \textsc{TimelyChat} benchmark to evaluate two key aspects: response timing and time-conditioned responses.
Additionally, we construct a large-scale training dataset by leveraging unlabeled event knowledge from a temporal commonsense knowledge graph and using a Large Language Model (LLM) to synthesize 55K event-driven dialogues.
We then train \textsc{Timer}, a dialogue agent designed to proactively predict time intervals and generate timely responses that align with those intervals.
Experimental results show that \textsc{Timer} outperforms instruction-tuned LLMs and other time-aware baselines on response timing and time-specificity in both turn-level and dialogue-level evaluations.
We publicly release our data, model, and code.\footnote{Code: \url{https://github.com/sb-jang/timelychat}}
\end{abstract}

\section{Introduction}
\label{sec:intro}
Open-domain dialogue systems have introduced various aspects to improve human-likeness of their own, such as persona~\citep{zhang-etal-2018-personalizing,ahn-etal-2023-mpchat}, long-term memory~\citep{xu-etal-2022-beyond,xu-etal-2022-long-cr}, commonsense~\citep{zhou-etal-2021-commonsense,qin-etal-2021-timedial}, emotional support~\citep{rashkin-etal-2019-towards,liu-etal-2021-towards,zhang-etal-2024-escot}, and role-play~\citep{shao-etal-2023-character,li2023chatharuhirevivinganimecharacter}.
These efforts have led to the success of commercial chat services like Replika and Character AI, which have met the public's demand for social companion chatbots~\citep{CHATURVEDI2023122634,guingrich2023chatbots}.
The pursuit of human-like chatbots still remains important with the advances in Large Language Models (LLMs), along with the growing demand for AI agents capable of engaging in more natural and human-like ways.

\begin{figure}[t]
\centering
    \includegraphics[width=\columnwidth]{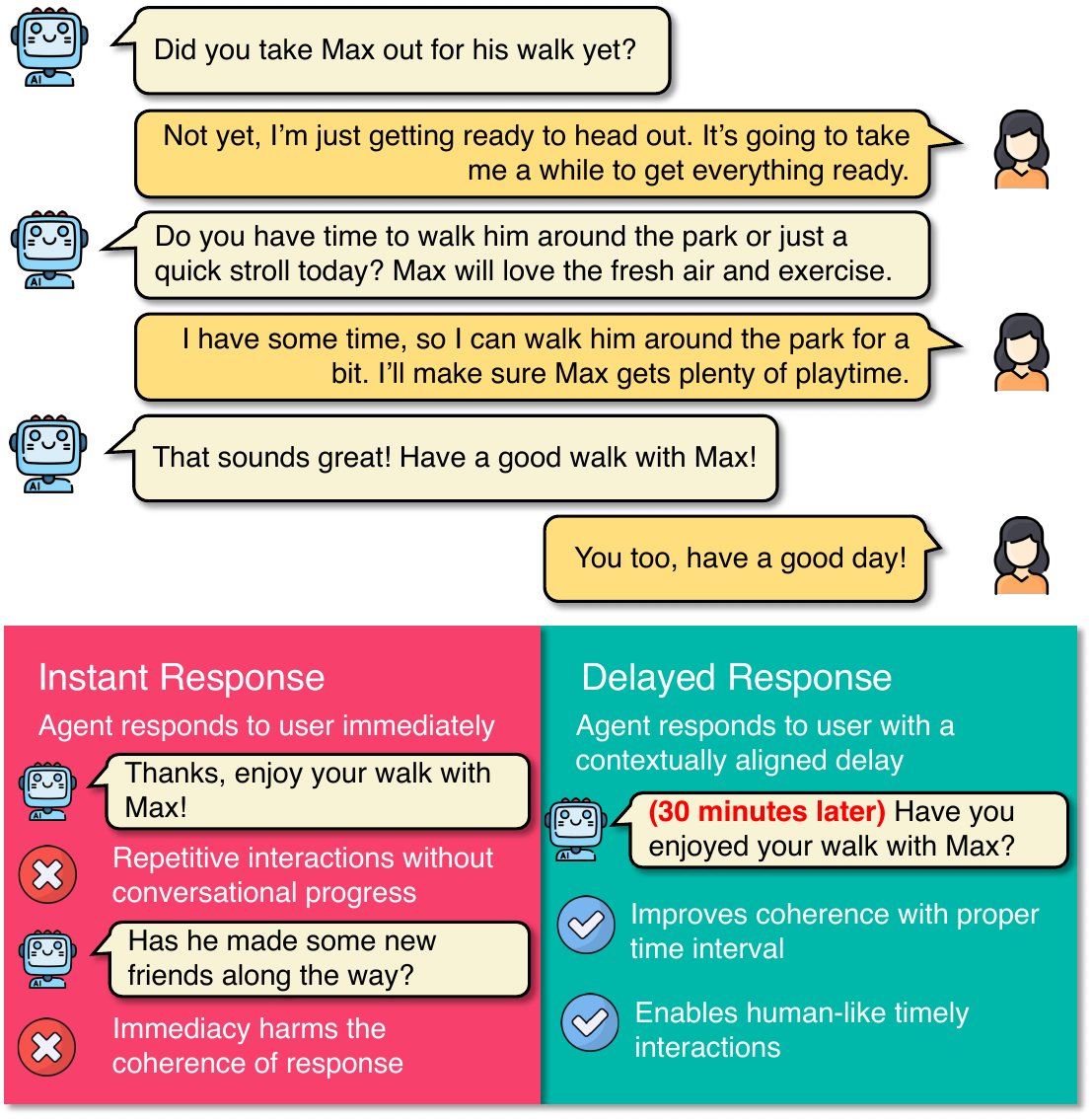}
    \caption{An illustrative example of a timely dialogue agent in an asynchronous messaging setting. Unlike a delay-agnostic agent that responds instantly with repetitive turns ignoring the ongoing event (walking the dog), a timely dialogue agent predicts an appropriate delay regime and re-engages with a response conditioned on that delay.}
    \label{fig:example}
\end{figure}

Research on dialogue response generation has predominantly focused on generating appropriate and consistent next utterances, conditioning on the textual information within dialogue contexts.
Meanwhile, although the question of \textit{what to respond} has received considerable attention, the problem of \textit{when to respond} remains underexplored.
For instance, as illustrated in Figure~\ref{fig:example}, if an agent generates only instant responses without considering response timing, it can cause repetitive interactions without conversational progress or produce responses that do not align with the temporal context of the conversational event.
In contrast, by incorporating response timing, an agent can maintain a natural flow while providing timely responses.
\textit{This requires grounding responses in the temporal context tied to the status of the event, mirroring the way humans naturally adapt their responses in human-to-human conversations.}
Across large-scale real messaging corpora, roughly 40\% of replies are delayed rather than instant (Appendix~\ref{appx:delay_prevalence}), yet existing dialogue systems implicitly assume instant turn-taking.
We stress that the goal is not to send a response at an exact predicted timestamp, but to decide an appropriate delay regime, i.e., whether a turn should be instant or delayed and roughly how long to wait, which then conditions the generated response.

Although \textit{when} to speak has been studied separately from \textit{what} to say in spoken dialogue~\citep{liao2025beyond}, the interval is given as a fixed input in text-based dialogue~\citep{zhang-etal-2023-mind}.
We invert this assumption and make the timing prediction the agent's first subtask.
Intervals in existing time-aware data are not predicted from the ongoing event, but assigned at session boundaries.
We thus propose both the task and a training method that supplies this supervision.

However, it is inherently challenging to simulate such scenarios with dialogue models trained on existing datasets.
Most dialogue datasets lack explicit temporal context.
Additionally, collecting real-time conversations where temporal context is naturally embedded (e.g., text messages between individuals) is highly restricted due to privacy concerns and ethical considerations.

In this work, we propose a novel task named \textbf{Timely Dialogue Response Generation}, which aims not only to generate coherent responses but also to consider the temporal context associated with ongoing events.
Specifically, it focuses on predicting the necessary time interval for the next utterance and generating a corresponding time-conditioned response.
We introduce the \textsc{TimelyChat} benchmark to assess both aspects.
The benchmark combines the \emph{human-annotated} event-duration pairs of MC-TACO~\citep{zhou-etal-2019-going-cr} with the dialogue generation capability of an LLM, and its 324 dialogues are manually screened.

Furthermore, we introduce a separate large-scale training set of 55K event-driven dialogues for supervised fine-tuning (SFT), built by the same recipe from a different source and generator.
We draw unlabeled events from the ATOMIC$_{20}^{20}$ knowledge graph~\citep{Hwang_Bhagavatula_Le_Bras_Da_Sakaguchi_Bosselut_Choi_2021} and have an LLM pseudo-label their durations.
This decouples the training and test labels, making the pipeline scalable without annotation and the benchmark independent of the training labels.
Using this dataset, we present \textsc{Timer}, a dialogue model fine-tuned with a multi-task learning objective that jointly predicts the time interval and generates the corresponding response.

Evaluation results on the proposed benchmark demonstrate that \textsc{Timer} outperforms both instruction-tuned LLMs and dialogue models trained on time-aware datasets in generating time-conditioned responses and predicting time intervals consistent with temporal commonsense.
Furthermore, in dialogue-level evaluations, \textsc{Timer} distinguishes between situations requiring delayed responses and those requiring instant responses more effectively, and generates more timely responses that align well with the predicted time intervals.

\section{Related Work}
\label{sec:related_work}

\paragraph{Long-term Dialogues.}
Long-term dialogue spans multiple sessions separated by time intervals.
Multi-Session Chat (MSC)~\citep{xu-etal-2022-beyond} introduces messaging-like multi-session chat. Conversation Chronicles (CC)~\citep{jang-etal-2023-conversation} adds a wider range of time intervals and fine-grained speaker relationships. LoCoMo~\citep{maharana-etal-2024-evaluating} extends to very long-term dialogue (up to 32 sessions) with a memory benchmark.
However, these focus on recalling personas or past events across sessions, rather than the temporal context between ongoing events in real-time conversations.
A notable exception, GapChat~\citep{zhang-etal-2023-mind}, introduces an event timeline to capture event progression over given time intervals.
Our work moves beyond predetermined intervals, requiring a proactive agent that dynamically determines realistic delays from temporal context.

\begin{figure*}[t]
\centering
    \includegraphics[width=\textwidth]{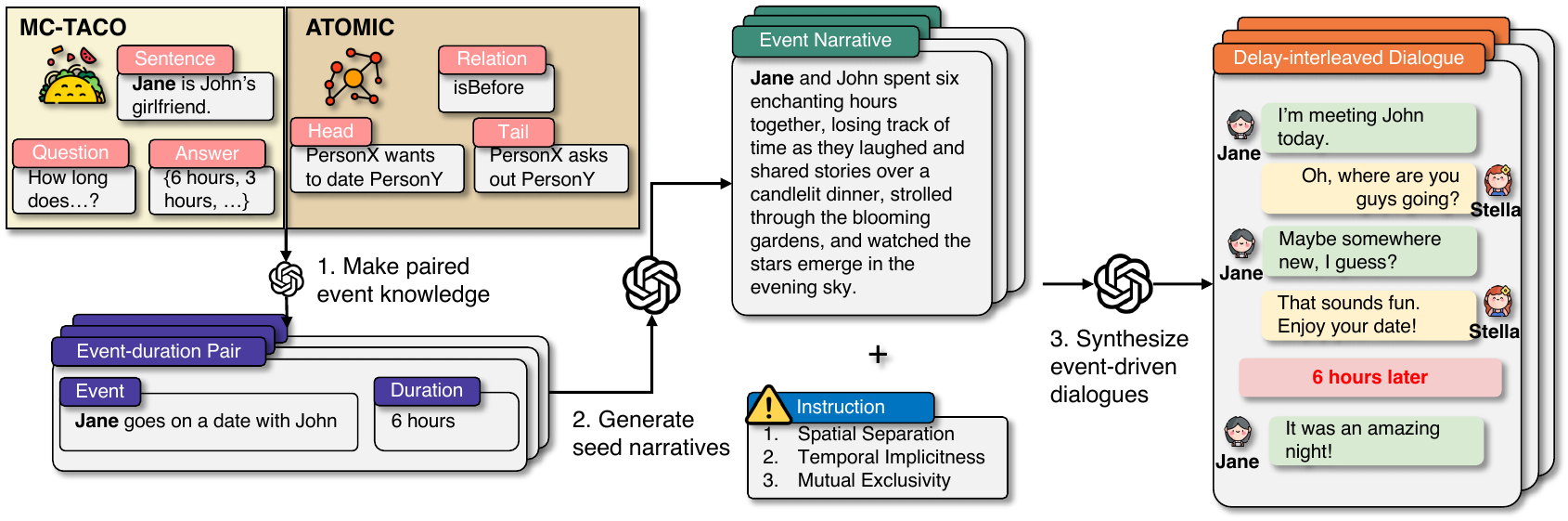}
    \caption{Overall process of the data construction method. Two different knowledge sources represent the same example for better understanding. Note that due to constraints imposed by the instruction, Jane's conversation partner becomes Stella, not John.}
    \label{fig:data_construction}
\end{figure*}

\paragraph{Proactive and Time-aware Dialogue Agents.}
Recent work increasingly treats response timing and proactivity as first-class problems.
One line models \emph{when} an agent should speak. \citet{liao2025beyond} cast turn-taking as a multimodal decision of when to speak in spoken interaction, while \citet{yang-etal-2025-mtpchat} introduce a time-aware persona dialogue benchmark with a temporal next response prediction task.
A parallel line builds proactive assistants that act without explicit user prompts~\citep{chen-etal-2025-llamapie,yang2025contextagent} and benchmarks to evaluate them~\citep{wang2025proactivevideoqa,yoon-etal-2025-beyond}, with \citet{wang2025proactivevideoqa} proposing a time-aware metric for proactive interactions.
Relatedly, \citet{cheng2026llmagentstemporallyblind} show that LLM agents are largely temporally blind, failing to account for the time elapsed between messages.
Unlike these efforts, which target spoken turn-taking, multimodal perception, or task assistance, we focus on text-based open-domain dialogue and jointly model both \emph{when} to respond and \emph{what} to respond within a single agent.

\section{Task Definition}
\label{sec:task}
We introduce a new task named \textbf{Timely Dialogue Response Generation}, which aims to generate contextually appropriate responses while incorporating temporal considerations from the dialogue history.
A key temporal factor that influences a response is how much time has passed since the previous utterance.
To capture this, we define \textit{time interval} as our primary temporal context, which represents the relative time difference (e.g., 10 minutes) between utterances.
Formally, we model the conditional probability distribution $P_\theta$ of a response $r_t$ at the $t$-th turn given the textual context $u$ and the temporal context $\tau$:
\begin{equation}
\label{eq:time-conditioned}
    r_t \sim P_\theta (u_t \vert u_1,\ldots,u_{t-1}, \tau_2,\ldots,\tau_t),
\end{equation}
where $u_t$ denotes the utterance at the $t$-th turn and $\tau_t$ denotes the elapsed time between $u_{t-1}$ and $u_t$.
This probability distribution can be further decomposed into two subtasks, which are the main focus of this work.

\paragraph{Subtask 1. Response Timing Prediction.}
The first task is to predict the optimal timing for delivering messages to users.
Mathematically, this involves predicting the $t$-th time interval $\hat{\tau}_t$ given the available contexts:
\begin{equation}
    \hat{\tau}_t \sim P_\theta (\tau_t \vert u_1,\ldots,u_{t-1}, \tau_2,\ldots,\tau_{t-1}). \label{eq:tdrg_time}
\end{equation}

\paragraph{Subtask 2. Time-conditioned Response Generation.}
The subsequent task is to generate a contextually appropriate response while incorporating the predicted timing for message delivery:
\begin{equation}
    r_t \sim P_\theta (u_t \vert u_1,\ldots,u_{t-1}, \tau_2,\ldots,\tau_{t-1}, \hat{\tau}_t). \label{eq:tdrg_response}
\end{equation}

Note that this task formulation challenges the widely held assumption that dialogue agents should always respond to user messages instantly.
Instead, it takes temporal context into account, i.e., the amount of elapsed time, to determine when a response should be generated.

\section{\textsc{TimelyChat} Benchmark}
\label{sec:timelychat}
We construct the \textsc{TimelyChat} benchmark to assess the timely response generation capabilities of dialogue models.
To this end, we first craft timely conversations through a temporal knowledge base and LLMs under explicit construction conditions, and then design two evaluation processes.
Figure~\ref{fig:data_construction} shows the overall construction process of our benchmark.
We release \textsc{TimelyChat} in two parts built by the same recipe: a high-quality benchmark of 324 manually screened test dialogues (this section) and a large-scale 55K training set (Section~\ref{sec:data_augmentation}), differing only in event source and generator.

\subsection{Data Construction}
\label{sec:data_construction}
We incorporate temporal information into dialogues using a temporal commonsense knowledge base.
This knowledge base captures various event-related temporal dynamics, and thus is well suited for transforming temporal context into event-driven dialogues.
By identifying temporal patterns, we seamlessly integrate them into conversations, using the dialogue generation capabilities of LLMs.

\paragraph{Event Knowledge Extraction.}
We first obtain a rich and reliable source of daily events and their typical durations for crafting event-driven conversations with temporal context.
To this end, we use the event-duration category of the MC-TACO dataset~\citep{zhou-etal-2019-going-cr}.
The dataset consists of sentences for specific events, queries to ask the typical duration of the event (e.g., ``How long does it take to \ldots?''), and human-annotated ground-truth answers.
We use the sentences with ground-truth answers, i.e., event-duration pairs, to synthesize event-driven conversations.
During data construction, we exclude the examples whose temporal intervals are shorter than one minute or longer than 24 hours to simulate realistic temporal delay in daily dialogue situations.
We instruct GPT-4o~\citep{achiam2023gpt} with these sentences and event-duration pairs to generate descriptive sentences.
It integrates the event and its duration into coherent sentences, forming seed narratives for dialogue generation.

\paragraph{Timely Dialogue Generation.}
\label{sec:timely_dialog_generation}
With the extracted temporal event knowledge, we instruct GPT-4o to generate conversations.
Our instruction contains the conditions that the generated dialogues must satisfy:
\begin{itemize}
    \item \textbf{Spatial Separation}: The scenario must involve one speaker experiencing an event while conversing with another speaker about it. This condition reproduces an asynchronous messaging setting rather than a conversation held where the event takes place, and prevents the contradiction of forcing a delay into a conversation that does not need one.
    \item \textbf{Temporal Implicitness}: The response must avoid direct references to the elapsed time. This condition reduces the occurrence of dull responses that simply acknowledge the time interval and, more importantly, prevents lexical overlap with the ground-truth time interval.
    It makes the interval inferable from the event rather than readable from the response text, and keeps the benchmark from rewarding a model that copies the interval instead of inferring it.
    \item \textbf{Mutual Exclusivity}: The time-conditioned response must become untimely under contrary temporal conditions. In other words, a delayed response should be incoherent under an instant condition with no time interval, and an instant response should be incoherent when a time interval exists. It prevents generating time-agnostic responses that remain coherent regardless of the temporal context.
\end{itemize}

Along with these instructions, we provide GPT-4o with one in-context demonstration, randomly sampled from six seed dialogues that the authors wrote by hand, each ranging from 5 to 10 turns.
These seeds serve only to prevent ill-formed outputs and diversify dialogue lengths; they are not part of the released data, and we find the choice of demonstration among the six to have a negligible effect on output quality.
Every generated dialogue is then manually inspected against the three conditions above, and any dialogue violating one of them is discarded, leaving a benchmark of 324 dialogues with an average length of 7.5 turns.
Its event coverage is inherited from the event-duration category of MC-TACO.
All prompts and examples used in the construction process are provided in Appendix~\ref{appx:data_construction_details}.

\begin{table*}[t]
\centering
\resizebox{\textwidth}{!}{%
\begin{tabular}{@{}lcccccc@{}}
\toprule
Dataset & \# Sessions & Construction Method & Time Granularity & Event-grounded & Timing Target & \# Events \\ \midrule
MSC~\citep{xu-etal-2022-beyond} & 13K (4.4K) & Crowdsourcing & hours--weeks & \xmark & \xmark & - \\
CC~\citep{jang-etal-2023-conversation} & 1M (160K) & LLM-generated & hours--years & \xmark & \xmark & - \\
LoCoMo~\citep{maharana-etal-2024-evaluating} & 842 (-) & LLM-gen + Crowd & days--months & \xmark & \xmark & - \\
GapChat~\citep{zhang-etal-2023-mind} & 2.6K (782) & LLM-gen + Crowd & minutes--years & \cmark & \xmark & 128 \\
\textbf{\textsc{TimelyChat} (Ours)} & 55K & LLM-generated & minutes--hours & \cmark & \cmark & 55K \\ \bottomrule
\end{tabular}%
}
\caption{Comparison of long-term dialogue datasets interleaved with time intervals. The number in parentheses under the \# Sessions column represents the count of sessions with time intervals within a day. Event-grounded indicates whether the dialogues reflect the temporal context associated with events or not. Timing Target indicates whether the response interval is supplied as a supervised prediction target rather than as a given input the model conditions on. For \textsc{TimelyChat}, \# Sessions refers to the 55K training dialogues. The benchmark additionally provides 324 manually screened test dialogues.}
\label{tab:stats}
\end{table*}

\subsection{Evaluation Protocols}
\label{sec:evaluation_tasks}
With the crafted conversations, we propose two evaluation approaches to assess the abilities of dialogue agents to generate timely responses: turn-level and dialogue-level.

\paragraph{Turn-level Evaluation.}
In turn-level evaluation, we assess each subtask on the target response.
For response timing prediction, a model predicts the time interval required for the next utterance given a dialogue context.
We then evaluate (1) whether the model correctly classifies the next turn as either delayed or instant, and (2) how close the predicted interval is to the ground truth.
We measure precision, recall, false positive rate (FPR), and F1 for the binary classification, and root mean squared logarithmic error (RMSLE) for regression by converting each time interval into minutes.
For response generation, a model generates a time-conditioned response given a dialogue context and ground-truth time interval.
We measure BLEU~\citep{papineni-etal-2002-bleu}, ROUGE~\citep{lin-2004-rouge}, and BERTScore~\citep{DBLP:conf/iclr/ZhangKWWA20} as reference-based metrics.
Additionally, we measure naturalness~\citep{mehri2022reportnsffuturedirections} and time-specificity~\citep{tsunomori-etal-2023-time} on a 5-point scale, adopting LLM-as-a-judge~\citep{liu-etal-2023-geval} for automatic evaluation.

\paragraph{Dialogue-level Evaluation.}
\label{sec:dialog-level_evaluation}
One crucial quality of a timely dialogue agent is its ability to introduce appropriate delays considering the temporal context while maintaining a natural conversational flow.
Inspired by dialogue-level evaluation methods with model-to-model interactions~\citep{li2019acuteevalimproveddialogueevaluation,zhou2024sotopia}, we provide an event-driven scenario and let an agent converse with GPT-4o as a user simulator~\citep{yoon-etal-2024-evaluating,kazi2024largelanguagemodelsuseragents,niu-etal-2024-enhancing} for a fixed number of turns to measure dialogue-level metrics.
We measure coherence~\citep{mehri2022reportnsffuturedirections} and dialogue-level time-specificity to assess the quality of the agent's responses, and measure delay-appropriateness that considers both the timing and duration of delays, using LLM-as-a-judge with a 5-point scale.
The evaluation criteria of LLM-as-a-judge metrics and simulator instructions are detailed in Appendix~\ref{appx:evaluation_setup_details}.

\section{\textsc{Timer}: A Dialogue Agent for Timely Responses}
\label{sec:timer}
\subsection{Training Data Augmentation with Unlabeled Knowledge}
\label{sec:data_augmentation}

To scale beyond the 324-dialogue benchmark, we construct a large training set using the same recipe and dialogue conditions as in Section~\ref{sec:data_construction}, changing only the inputs to favor scale over manual curation.

\paragraph{Knowledge Source and Durations.}
Instead of MC-TACO, we draw events from the ATOMIC$_{20}^{20}$ knowledge graph~\citep{Hwang_Bhagavatula_Le_Bras_Da_Sakaguchi_Bosselut_Choi_2021}, which provides far more event triplets but no annotated durations.
We prompt GPT-3.5 to convert each triplet into a single-sentence event description, replacing anonymized names (e.g., PersonX) with common names following~\citet{kim-etal-2023-soda}, and to estimate the event's typical duration as a pseudo-label, discarding durations under a minute or over 24 hours.

\paragraph{Generation at Scale.}
We then generate dialogues with GPT-3.5 (rather than GPT-4o) for cost efficiency, using the same instructions and conditions as the benchmark.
To curb ill-formed outputs (e.g., speaker mismatches) without manual screening, each prompt includes a one-shot demonstration randomly sampled from the benchmark.
The resulting set comprises 55K examples as shown in Table~\ref{tab:stats}, offering far more event-grounded dialogues at finer time granularity than prior datasets, without costly human annotation.
To measure its quality, two authors independently assess 200 random samples from the validation split and obtain per-condition pass rates of 92.8--100\% at $\kappa=0.918$ (Appendix~\ref{appx:trainset_audit}).
All prompts are provided in Appendix~\ref{appx:chatgpt_prompts}.

\subsection{Time-augmented Training with Multi-task Learning Objectives}
\label{sec:training_method}

Our training approach predicts an appropriate time interval for delaying the response based on the temporal context and then generates a time-conditioned response, introducing a time interval prediction step before each turn's utterance (Eqs.~\ref{eq:tdrg_time} and~\ref{eq:tdrg_response}).
For each turn consisting of a speaker identifier and a text utterance, we insert a time interval.
We prepend prefix tokens to distinguish each component, formatting the input as \texttt{<SPK> $s_t$ <TIME> $\tau_t$ <UTT> $u_t$}, where $s_t, \tau_t$, and $u_t$ denote the speaker, the time interval, and the utterance at the $t$-th turn, respectively.
For turns within the dialogue context, we set $\tau=0$, indicating no delay, to maintain coherence and align with typical instant responses.
$\hat{\tau}_t$ is the token sequence decoded after the \texttt{<TIME>} prefix, a number and a unit generated by beam search on the model's output distribution.

From these inputs, we define two losses for training: response timing prediction loss and response generation loss.
The losses are defined as follows:
\begin{equation}
\begin{split}
    \mathcal{L}_{\text{time}} &= -\frac{1}{N} \sum_{i=1}^N \sum_{t=2}^T \log p(\tau_t \mid s_{\leq t}, \tau_{<t}, u_{<t}), \\
    \mathcal{L}_{\text{response}} &= -\frac{1}{N} \sum_{i=1}^N \sum_{t=2}^T \log p(u_t \mid s_{\leq t}, \tau_{\leq t}, u_{<t}),
\end{split}
\end{equation}
where $N$ is the number of training examples, and $T$ is the number of turns in a dialogue.

The final multi-task learning objective is given as follows:
\begin{align}
\label{eq:mt_loss}
    \mathcal{L} = \mathcal{L}_{\text{response}} + \lambda \mathcal{L}_{\text{time}}.
\end{align}
This approach ensures that the model learns both to predict appropriate time intervals and to generate time-conditioned responses effectively.

\section{Experiments}
\label{sec:experiments}
\subsection{Baselines}

We evaluate two types of dialogue agents for simulating timely dialogue response generation: instruction-tuned models and time-aware models.
The instruction-tuned models include LLMs optimized for dialogue use cases.
We select 8B and 70B models of LLaMA 3.1 Instruct~\citep{dubey2024llama} as open-source chat models, and GPT-3.5 and GPT-4o as proprietary models.
We experiment with zero-shot, few-shot, and chain-of-thought (CoT)~\citep{10.5555/3600270.3602070} prompting strategies to investigate the effectiveness of in-context learning without task-specific fine-tuning.
The time-aware models are trained on dialogue datasets where time intervals are interleaved.
We compare the following models:

\begin{itemize}
    \item \textbf{MSC 3B}~\citep{xu-etal-2022-beyond}: Trained on BlenderBot~\citep{roller-etal-2021-recipes} using the MSC dataset, which includes time intervals between sessions.
    \item \textbf{ReBot 400M}~\citep{jang-etal-2023-conversation}: Trained on BART-Large~\citep{lewis-etal-2020-bart} using the CC dataset, which consists of large-scale LLM-generated dialogues.
    \item \textbf{GapChat 3B}~\citep{zhang-etal-2023-mind}: Trained on MSC using the GapChat dataset, which incorporates event progress based on time intervals.
\end{itemize}

To prevent a judge LLM from introducing potential biases such as self-preference bias~\citep{wataoka2024self,panickssery2024llm}, we use Claude Sonnet 4.5 as the evaluator across both turn-level and dialogue-level evaluations.
Implementation details of all models including \textsc{Timer} 3B are described in Appendix~\ref{appx:implementation_details}.

\begin{table}[t]
\centering
\resizebox{\columnwidth}{!}{%
\begin{tabular}{@{}lccccc@{}}
\toprule
Model & Precision $\uparrow$ & Recall $\uparrow$ & F1$\uparrow$ & FPR$\downarrow$ & RMSLE$\downarrow$ \\ \midrule
\textbf{LLaMA 3.1 8B} &  &  &  &  &  \\
\textit{Zero-shot} & 0.172 & 0.691 & 0.276 & 0.606 & 2.853 \\
\textit{Few-shot} & 0.164 & \textbf{0.920} & 0.279 & 0.855 & 2.807 \\
\textit{CoT} & 0.160 & 0.824 & 0.268 & 0.790 & 2.854 \\
\textbf{LLaMA 3.1 70B} &  &  &  &  &  \\
\textit{Zero-shot} & 0.153 & 0.735 & 0.254 & 0.740 & 2.479 \\
\textit{Few-shot} & 0.197 & 0.602 & 0.297 & 0.448 & 2.326 \\
\textit{CoT} & 0.194 & 0.836 & 0.315 & 0.636 & 3.066 \\
\textbf{GPT-3.5} &  &  &  &  &  \\
\textit{Zero-shot} & 0.143 & 0.784 & 0.241 & 0.861 & 2.763 \\
\textit{Few-shot} & 0.212 & 0.349 & 0.264 & 0.237 & 2.146 \\
\textit{CoT} & 0.186 & 0.762 & 0.299 & 0.609 & 2.667 \\
\textbf{GPT-4o} &  &  &  &  &  \\
\textit{Zero-shot} & 0.266 & 0.259 & 0.263 & 0.131 & 1.956 \\
\textit{Few-shot} & 0.227 & 0.423 & 0.295 & 0.099 & 2.252 \\
\textit{CoT} & 0.202 & 0.852 & 0.326 & 0.615 & 2.938 \\ \midrule
\textsc{Timer} 3B (Ours) & \textbf{0.783} & 0.799 & \textbf{0.791} & \textbf{0.041} & \textbf{1.189} \\ \bottomrule
\end{tabular}
}
\caption{Results of response timing prediction. For few-shot and CoT strategies, we provide balanced 2-shot demonstrations which consist of one delayed example and one instant example, along with the task description used in zero-shot prompting.}
\label{tab:response_timing_prediction}
\end{table}

\begin{figure*}[t]
\centering
    \includegraphics[width=\textwidth]{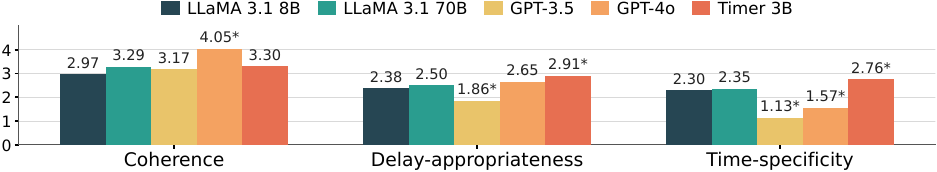}
    \caption{Results of simulated interactions for timely dialogue agents. We perform pairwise $t$-tests and denote statistically significant score differences from the other models with an asterisk (*) ($p<0.05$).}
    \label{fig:dialog-level_results}
\end{figure*}

\subsection{Turn-level Evaluation Results}

\paragraph{Response Timing Prediction.}
Table~\ref{tab:response_timing_prediction} presents the results of response timing prediction on the \textsc{TimelyChat} benchmark.
Overall, instruction-tuned models show low precision and F1 with high FPR, indicating that they over-predict delays and introduce unnecessary intervals that disrupt the conversational flow.
Although few-shot and CoT strategies slightly improve F1 scores across all LLMs, they sometimes negatively impact FPR compared to zero-shot prompting.
In contrast, \textsc{Timer} 3B achieves the highest F1 score and the lowest FPR compared to instruction-tuned models.
Even the best-performing GPT-4o, which generated the benchmark, still lags significantly behind the fine-tuned \textsc{Timer} 3B model.

Likewise, when it comes to predicting the length of time intervals, in-context learning methods fail to enhance performance effectively.
While few-shot prompting achieves a lower RMSLE than CoT across all LLMs, it does not consistently outperform zero-shot prompting, as demonstrated by GPT-4o's results.
These findings indicate that prompting with task descriptions and demonstrations alone is insufficient to reliably predict whether to pose a delay and how long it should last.
In contrast, task-specific fine-tuning is essential for effectively learning these capabilities.

\begin{table}[t]
\centering
\resizebox{\columnwidth}{!}{%
\begin{tabular}{@{}lcccccc@{}}
\toprule
Model & B-2 & R-L & BS & Nat. & Spec. \\ \midrule
\rowcolor{gray!20} \multicolumn{6}{c}{\textsc{Instruction-tuned Models}} \\
\textbf{LLaMA 3.1 8B} &  &  &  &  & \\
\textit{Zero-shot} & 5.38 & 12.38 & 86.21 & 2.24 & 2.78 \\
\textit{Few-shot} & 7.63 & 13.47 & 86.85 & 2.81 & 2.83 \\
\textit{CoT} & 6.17 & 12.50 & 86.23 & 1.31 & 1.84 \\
\textbf{LLaMA 3.1 70B} &  &  &  &  & \\
\textit{Zero-shot} & 6.84 & 12.71 & 85.90 & 2.86 & 2.35  \\
\textit{Few-shot} & 8.35 & 14.83 & 87.16 & 3.92 & 2.89 \\
\textit{CoT} & 9.01 & 15.01 & 87.12 & 1.07 & 3.04 \\
\textbf{GPT-3.5} &  &  &  &  & \\
\textit{Zero-shot} & 9.97 & 17.13 & 87.54 & 4.26 & 2.86 \\
\textit{Few-shot} & 11.23 & 17.81 & 87.77 & 4.25 & 2.76 \\
\textit{CoT} & 8.86 & 15.14 & 86.79 & 2.38 & 2.15 \\
\textbf{GPT-4o} &  &  &  &  & \\
\textit{Zero-shot} & 9.17 & 16.76 & 87.35 & 4.75 & 3.09 \\
\textit{Few-shot} & 11.15 & 18.51 & 87.91 & \textbf{4.78} & 3.09 \\
\textit{CoT} & 10.25 & 17.13 & 87.52 & 3.21 & 2.84 \\ \midrule
\rowcolor{gray!20} \multicolumn{6}{c}{\textsc{Time-aware Models}} \\
MSC 3B & 3.26 & 8.94 & 85.18 & 1.36 & 1.02 \\
ReBot 400M & 3.85 & 9.32 & 85.72 & 3.26 & 1.02 \\
GapChat 3B & 3.59 & 8.61 & 85.22 & 1.87 & 1.05 \\
\textsc{Timer} 3B (Ours) & \textbf{16.08} & \textbf{22.26} & \textbf{88.74} & 3.84 & \textbf{3.36} \\ \bottomrule
\end{tabular}%
}
\caption{Results of time-conditioned response generation on \textsc{TimelyChat}. B-2, R-L, BS, Nat., and Spec. refer to BLEU-2, ROUGE-L, BERTScore, naturalness, and time-specificity, respectively.}
\label{tab:time-conditioned_response_generation}
\end{table}

\paragraph{Time-conditioned Response Generation.}
Table~\ref{tab:time-conditioned_response_generation} shows the time-conditioned response generation performance on the \textsc{TimelyChat} benchmark.
For instruction-tuned models, we observe that zero-shot performance tends to improve as model size increases across all metrics.
Among all LLMs, few-shot prompting consistently outperforms zero-shot prompting, while CoT prompting performs the worst in terms of naturalness and time-specificity.
This aligns with previous findings that LLMs struggle to generate helpful CoT rationales for dialogue response generation~\citep{chae-etal-2023-dialogue}.

Meanwhile, models trained on existing dialogue datasets that include time intervals exhibit poor overall performance.
Notably, these models achieve low time-specificity, indicating that they struggle to generate timely responses conditioned on given time intervals.
This stems from the characteristics that time intervals in existing long-term dialogue datasets are assigned arbitrarily rather than based on the temporal context of ongoing events, making it difficult for models to learn the conditional distribution of responses based on the given interval.
For example, we find that these models frequently generate generic greeting messages, failing to capture the temporal nuances of timely responses.
In contrast, \textsc{Timer} 3B, despite having a smaller model size, achieves comparable naturalness to instruction-tuned LLMs and even surpasses LLaMA 3.1 models.
More importantly, it achieves the highest time-specificity, demonstrating that our training approach enables response generation that aligns well with event-specific temporal contexts.

\subsection{Dialogue-level Evaluation Results}

Beyond turn-level evaluation, we also conduct dialogue-level evaluation to assess whether a dialogue agent can introduce temporally contextual delays at appropriate timing without disrupting the conversational flow.
We let the four zero-shot LLMs from the previous experiments, along with \textsc{Timer} 3B, engage in 10 interactions with the simulator described in Section~\ref{sec:evaluation_tasks}.
To simulate event-driven dialogue, we provide the first turn of conversations from the \textsc{TimelyChat} benchmark as the initial interaction.

We sample dialogues that include at least one delayed response and report three dialogue-level metrics in Figure~\ref{fig:dialog-level_results}.
GPT-4o achieves the highest coherence among the models, demonstrating its ability to maintain a natural conversation flow, while \textsc{Timer} 3B achieves the second-highest coherence score.
Notably, \textsc{Timer} 3B shows the highest delay-appropriateness and time-specificity scores.
This suggests that \textsc{Timer} 3B effectively considers both dialogue context and temporal context to predict delays with appropriate timing and duration.
Additionally, it generates delayed responses that are coherent only when a delay is given, thereby justifying and necessitating the delay.
In contrast, LLaMA 3.1 8B and 70B exhibit relatively lower delay-appropriateness, while GPT-3.5 and GPT-4o achieve lower time-specificity scores.
We further analyze these findings in the case study in Appendix~\ref{appx:case_study}.
The result shows that \textsc{Timer} 3B introduces a delay when the user's utterance suggests a natural pause and justifies it in the response, whereas LLaMA 3.1 70B delays in every turn with arbitrary intervals and GPT-4o predicts plausible intervals but fails to generate time-specific responses that justify them.

\begin{table}[t]
\centering
\scriptsize
\setlength{\tabcolsep}{4pt}
\resizebox{\columnwidth}{!}{%
\begin{tabular}{@{}llccccc@{}}
\toprule
Setting & Metric & Win & Tie & Loss & $\kappa$ & $p$ \\ \midrule
\multirow{2}{*}{Turn-level}     & Naturalness   & 19\%          & 24\% & \textbf{57\%} & 0.446 & $<0.001$ \\
                                & Time-spec.    & \textbf{54\%} & 24\% & 22\%          & 0.508 & $<0.001$ \\ \midrule
\multirow{3}{*}{Dialogue-level} & Coherence     & 21\%          & 18\% & \textbf{61\%} & 0.637 & $<0.001$ \\
                                & Delay-appr.   & \textbf{40\%} & 23\% & 37\%          & 0.446 & $0.899$ \\
                                & Time-spec.    & \textbf{40\%} & 23\% & 37\%          & 0.496 & $0.899$ \\ \bottomrule
\end{tabular}%
}
\caption{Pairwise human evaluation results on both turn-level and dialogue-level metrics. Win/Tie/Loss rates of \textsc{Timer} 3B against zero-shot GPT-4o are presented. Delay-appr.\ and Time-spec.\ denote delay-appropriateness and time-specificity. Fleiss' $\kappa$ is also reported to measure inter-annotator agreement. $p$ is from a two-sided exact sign test over the decisive comparisons, excluding ties.}
\label{tab:human_eval}
\end{table}

\subsection{Human Evaluation Results}

To investigate the reliability of LLM-based evaluation, we also conduct human evaluations on both turn-level and dialogue-level metrics.
We recruit twenty graduate students as annotators, who are fluent in English but have no background in our work.
We provide them with the same evaluation criteria used for LLM-based evaluation, and ask them to perform blinded comparisons between responses or dialogues from two different models.
Each example is evaluated three times by different annotators, and aggregated by majority voting.
Table~\ref{tab:human_eval} presents the results for both turn-level and dialogue-level metrics.

For turn-level metrics, we compare \textsc{Timer} 3B with the most competitive baseline, zero-shot GPT-4o, on 200 randomly sampled examples.
While \textsc{Timer} 3B falls short of GPT-4o in terms of naturalness, it outperforms GPT-4o in time-specificity, and this margin is significant under a two-sided sign test.
This is consistent with the LLM-based evaluation results observed in Table~\ref{tab:time-conditioned_response_generation}.

For dialogue-level metrics, we compare \textsc{Timer} 3B with zero-shot GPT-4o on 80 randomly sampled examples.
Again, \textsc{Timer} 3B lags behind GPT-4o in coherence, but it achieves comparable or better results on both delay-appropriateness and time-specificity.
These two margins are not significant under the same test.
Nevertheless, their direction aligns with the results shown in Figure~\ref{fig:dialog-level_results}, indicating that human judgment agrees with our LLM-based evaluation against the strongest baseline.
This suggests that \textsc{Timer} 3B is specialized in temporal control, while its gaps in naturalness and coherence remain significant.
We observe that the annotators show substantial ($0.6<\kappa\leq0.8$) agreement on coherence, and moderate ($0.4<\kappa\leq0.6$) agreement on the other metrics, which is in the range commonly reported for subjective dialogue-quality judgments.
To further verify that our LLM-based evaluation is not an artifact of a single judge, we additionally evaluate the same samples with another judge model (GPT-4o) and report inter-judge consistency in Appendix~\ref{appx:reliability}.

\begin{table}[t]
\centering
\resizebox{\columnwidth}{!}{%
\begin{tabular}{@{}lccccc@{}}
\toprule
\multirow{2}{*}{Method} & \multicolumn{2}{c}{Timing Prediction}   & \multicolumn{3}{c}{Response Generation} \\ \cmidrule(lr){2-3} \cmidrule(lr){4-6}
                   & F1     & RMSLE & R-L   & Nat. & Spec. \\ \midrule
\textbf{\textsc{Timer} 3B}       & \textbf{0.791} & \textbf{1.189} & \textbf{22.26} & \textbf{3.84} & \textbf{3.36} \\
w/o Special Tokens & 0.299 & 3.700 & 15.25 & 1.39 & 2.37  \\
Utterance First       & 0.249 & 3.451 & 12.07 & 1.57 & 1.81  \\
w/o Time Loss         & 0.299 & 3.987 & 13.38 & 2.54 & 2.43  \\ \bottomrule
\end{tabular}%
}
\caption{Ablation study on each component of our training scheme. In the Utterance First setting, we train a model to generate an utterance first and then predict the response timing conditioned on it.}%
\label{tab:ablation}
\end{table}

\subsection{Ablation Study}

To demonstrate the effectiveness of the proposed training scheme, we provide the impacts of each component on task performance in Table~\ref{tab:ablation}.

\paragraph{Effects of Input Formats.}
Special tokens (e.g., \texttt{<UTT>}, \texttt{<TIME>}) play a crucial role in indicating which component of an utterance to generate.
Although trained with both time intervals and responses, the model trained without them struggles to predict time intervals at the right place, yielding low F1 and high RMSLE.
This results in arbitrary delayed or instant responses, causing poor controllability and also degrading response generation quality.

The order of performing the two subtasks significantly affects performance.
When trained to generate a response first and then predict the interval conditioned on it, the model tends to predict delays even for instant responses, resulting in poor timing prediction.
This format also yields the lowest ROUGE-L among the settings, since the model is not provided with time conditions directly for generation.

\paragraph{Effects of Timing Prediction Loss.}
We investigate the effectiveness of our multi-task objective.
When the timing prediction loss is removed ($\lambda=0$), the model struggles to learn timing prediction itself, resulting in random predictions.
Notably, response generation performance degrades compared to \textsc{Timer} 3B, even though the model is still optimized with time-conditioned context inputs.
This implies that the timing prediction objective helps improve response generation capability, demonstrating the necessity of the proposed training objective.

\section{Conclusion}
\label{sec:conclusion}
We highlight that open-domain dialogue agents must decide not only what to say but when, grounded in the temporal context of the ongoing event.
We formulate this as timely dialogue response generation and introduce the \textsc{TimelyChat} benchmark for turn- and dialogue-level evaluation.
We also train \textsc{Timer} 3B on a 55K training set with a time-augmented objective, so that it predicts an interval and then generates a response conditioned on it.
\textsc{Timer} 3B outperforms baselines on timing, while it still falls short of GPT-4o in naturalness and coherence.

\section{Limitations}
\label{sec:limitations}

\paragraph{Synthetic Data and Real-world Grounding.}

Both our benchmark and training set are synthetic, and we present no validation on real conversations.
Our real-corpus analysis (Appendix~\ref{appx:delay_prevalence}) shows that delayed responding is pervasive but does not isolate the event-grounded delays we target, which coexist with unavailability, platform characteristics, and social norms.
Since this subset cannot be separated from raw logs by mechanical filtering, we synthesize it under controlled constraints.
Cross-dataset evaluation is not directly available either, because existing time-aware dialogue benchmarks supply the interval instead of supervising its prediction.
We leave verification on anonymized real-world dialogue, including how often delays there are event-grounded, to future work.

\paragraph{Scope and Time Granularity.}
Our scope is intentionally event-driven, where temporal context most directly governs timing, so generalization to non-event settings remains open.
We also model event duration as discrete values (e.g., 30 minutes) rather than continuous ranges (e.g., 2--6 hours), and interval or distributional prediction is a natural extension for finer-grained control.

\paragraph{Model and Evaluation.}
\textsc{Timer} 3B is strongest on time-specificity and delay-appropriateness but lags behind the much larger GPT-4o in naturalness and coherence, and response timing is only an observable proxy for an interlocutor's latent state.
Fluency-oriented objectives (e.g., LLM-judge rewards) and larger model scales are promising next steps.
Finally, our dialogue-level evaluation relies on short simulated dialogues, and longer, more diverse interactions could further strengthen the analysis.

\section{Ethics Statement}
\label{sec:ethics}
The proposed dataset was designed to assess capabilities related to response timing and time-conditioned response in event-driven conversations.
To achieve this, we used event knowledge from publicly available datasets from various sources, and LLM-generated content either without or with some modification if necessary.
During this process, there is a possibility that harmful content or inappropriate biases existing in the original data may have been conveyed, or may have arisen due to limitations of filtering techniques.
We reject any form of violence, discrimination, or offensive language, and our dataset and experimental results do not represent such values.
If any harmful content or privacy infringement is identified within our data, we kindly request immediate notification to the authors.
In the event of such cases being reported, we will apply the highest ethical standards and take appropriate actions.

We release our code under the MIT License, our model under Apache 2.0, and our data under CC BY.
The benchmark events are drawn from MC-TACO~\citep{zhou-etal-2019-going-cr} and the training events from the ATOMIC$_{20}^{20}$ knowledge graph~\citep{Hwang_Bhagavatula_Le_Bras_Da_Sakaguchi_Bosselut_Choi_2021}, which is distributed under CC BY, and the dialogues are synthesized with OpenAI models in accordance with their terms of use.
We use each of these resources only for research.

\section*{Acknowledgments}
We thank the anonymous reviewers for their constructive comments.
This work was supported by the National Research Foundation of Korea (NRF) grant funded by the Korea government (MSIT) (No.~RS-2024-00335873).

\bibliography{anthology-1,anthology-2,custom}

\begin{thebibliography}{47}
\providecommand{\natexlab}[1]{#1}

\bibitem[{Achiam et~al.(2023)Achiam, Adler, Agarwal, Ahmad, Akkaya, Aleman,
  Almeida, Altenschmidt, Altman, Anadkat et~al.}]{achiam2023gpt}
Josh Achiam, Steven Adler, Sandhini Agarwal, Lama Ahmad, Ilge Akkaya,
  Florencia~Leoni Aleman, Diogo Almeida, Janko Altenschmidt, Sam Altman,
  Shyamal Anadkat, et~al. 2023.
\newblock \href {https://arxiv.org/abs/2303.08774} {{GPT-4} technical report}.
\newblock \emph{Preprint}, arXiv:2303.08774.

\bibitem[{Ahn et~al.(2023)Ahn, Song, Yun, and Kim}]{ahn-etal-2023-mpchat}
Jaewoo Ahn, Yeda Song, Sangdoo Yun, and Gunhee Kim. 2023.
\newblock \href {https://doi.org/10.18653/v1/2023.acl-long.189} {{MPCHAT}:
  Towards multimodal persona-grounded conversation}.
\newblock In \emph{Proceedings of the 61st Annual Meeting of the Association
  for Computational Linguistics (Volume 1: Long Papers)}, pages 3354--3377,
  Toronto, Canada. Association for Computational Linguistics.

\bibitem[{Chae et~al.(2023)Chae, Song, Ong, Kwon, Kim, Yu, Lee, Kang, and
  Yeo}]{chae-etal-2023-dialogue}
Hyungjoo Chae, Yongho Song, Kai Ong, Taeyoon Kwon, Minjin Kim, Youngjae Yu,
  Dongha Lee, Dongyeop Kang, and Jinyoung Yeo. 2023.
\newblock \href {https://doi.org/10.18653/v1/2023.emnlp-main.342} {Dialogue
  chain-of-thought distillation for commonsense-aware conversational agents}.
\newblock In \emph{Proceedings of the 2023 Conference on Empirical Methods in
  Natural Language Processing}, pages 5606--5632, Singapore. Association for
  Computational Linguistics.

\bibitem[{Chaturvedi et~al.(2023)Chaturvedi, Verma, Das, and
  Dwivedi}]{CHATURVEDI2023122634}
Rijul Chaturvedi, Sanjeev Verma, Ronnie Das, and Yogesh~K. Dwivedi. 2023.
\newblock \href {https://doi.org/10.1016/j.techfore.2023.122634} {Social
  companionship with artificial intelligence: Recent trends and future
  avenues}.
\newblock \emph{Technological Forecasting and Social Change}, 193:122634.

\bibitem[{Chen et~al.(2025)Chen, Batchelder, Liu, Smith, and
  Gollakota}]{chen-etal-2025-llamapie}
Tuochao Chen, Nicholas~Scott Batchelder, Alisa Liu, Noah~A. Smith, and
  Shyamnath Gollakota. 2025.
\newblock \href {https://doi.org/10.18653/v1/2025.findings-acl.710}
  {{L}lama{PIE}: Proactive in-ear conversation assistants}.
\newblock In \emph{Findings of the Association for Computational Linguistics:
  ACL 2025}, pages 13801--13824, Vienna, Austria. Association for Computational
  Linguistics.

\bibitem[{Cheng et~al.(2026)Cheng, Moakhar, Fan, Hosseini, Faghih, Sodagar,
  Wang, and Feizi}]{cheng2026llmagentstemporallyblind}
Yize Cheng, Arshia~Soltani Moakhar, Chenrui Fan, Parsa Hosseini, Kazem Faghih,
  Zahra Sodagar, Wenxiao Wang, and Soheil Feizi. 2026.
\newblock \href {https://doi.org/10.18653/v1/2026.findings-acl.1848} {Your
  {LLM} agents are temporally blind: The misalignment between tool use
  decisions and human time perception}.
\newblock In \emph{Findings of the {A}ssociation for {C}omputational
  {L}inguistics: {ACL} 2026}, pages 37082--37104, San Diego, California, United
  States. Association for Computational Linguistics.

\bibitem[{Dubey et~al.(2024)Dubey, Jauhri, Pandey, Kadian, Al-Dahle, Letman,
  Mathur, Schelten, Yang, Fan et~al.}]{dubey2024llama}
Abhimanyu Dubey, Abhinav Jauhri, Abhinav Pandey, Abhishek Kadian, Ahmad
  Al-Dahle, Aiesha Letman, Akhil Mathur, Alan Schelten, Amy Yang, Angela Fan,
  et~al. 2024.
\newblock \href {https://arxiv.org/abs/2407.21783} {The {Llama} 3 herd of
  models}.
\newblock \emph{Preprint}, arXiv:2407.21783.

\bibitem[{Guingrich and Graziano(2025)}]{guingrich2023chatbots}
Rose~E. Guingrich and Michael S.~A. Graziano. 2025.
\newblock \href {https://doi.org/10.1093/9780198945215.003.0011} {Chatbots as
  social companions}.
\newblock In \emph{Oxford Intersections: {AI} in Society}. Oxford University
  Press.

\bibitem[{Hwang et~al.(2021)Hwang, Bhagavatula, Le~Bras, Da, Sakaguchi,
  Bosselut, and
  Choi}]{Hwang_Bhagavatula_Le_Bras_Da_Sakaguchi_Bosselut_Choi_2021}
Jena~D. Hwang, Chandra Bhagavatula, Ronan Le~Bras, Jeff Da, Keisuke Sakaguchi,
  Antoine Bosselut, and Yejin Choi. 2021.
\newblock \href {https://doi.org/10.1609/aaai.v35i7.16792} {({Comet}-) {Atomic}
  2020: On symbolic and neural commonsense knowledge graphs}.
\newblock \emph{Proceedings of the AAAI Conference on Artificial Intelligence},
  35(7):6384--6392.

\bibitem[{Jang et~al.(2023)Jang, Boo, and Kim}]{jang-etal-2023-conversation}
Jihyoung Jang, Minseong Boo, and Hyounghun Kim. 2023.
\newblock \href {https://doi.org/10.18653/v1/2023.emnlp-main.838} {Conversation
  chronicles: Towards diverse temporal and relational dynamics in multi-session
  conversations}.
\newblock In \emph{Proceedings of the 2023 Conference on Empirical Methods in
  Natural Language Processing}, pages 13584--13606, Singapore. Association for
  Computational Linguistics.

\bibitem[{Kazi et~al.(2024)Kazi, Lyu, Zhou, Hakkani-Tur, and
  Tur}]{kazi2024largelanguagemodelsuseragents}
Taaha Kazi, Ruiliang Lyu, Sizhe Zhou, Dilek Hakkani-Tur, and Gokhan Tur. 2024.
\newblock \href {https://arxiv.org/abs/2411.09972} {Large language models as
  user-agents for evaluating task-oriented-dialogue systems}.
\newblock \emph{Preprint}, arXiv:2411.09972.

\bibitem[{Kim et~al.(2023)Kim, Hessel, Jiang, West, Lu, Yu, Zhou, Bras,
  Alikhani, Kim, Sap, and Choi}]{kim-etal-2023-soda}
Hyunwoo Kim, Jack Hessel, Liwei Jiang, Peter West, Ximing Lu, Youngjae Yu, Pei
  Zhou, Ronan Bras, Malihe Alikhani, Gunhee Kim, Maarten Sap, and Yejin Choi.
  2023.
\newblock \href {https://doi.org/10.18653/v1/2023.emnlp-main.799} {{SODA}:
  Million-scale dialogue distillation with social commonsense
  contextualization}.
\newblock In \emph{Proceedings of the 2023 Conference on Empirical Methods in
  Natural Language Processing}, pages 12930--12949, Singapore. Association for
  Computational Linguistics.

\bibitem[{Lewis et~al.(2020)Lewis, Liu, Goyal, Ghazvininejad, Mohamed, Levy,
  Stoyanov, and Zettlemoyer}]{lewis-etal-2020-bart}
Mike Lewis, Yinhan Liu, Naman Goyal, Marjan Ghazvininejad, Abdelrahman Mohamed,
  Omer Levy, Veselin Stoyanov, and Luke Zettlemoyer. 2020.
\newblock \href {https://doi.org/10.18653/v1/2020.acl-main.703} {{BART}:
  Denoising sequence-to-sequence pre-training for natural language generation,
  translation, and comprehension}.
\newblock In \emph{Proceedings of the 58th Annual Meeting of the Association
  for Computational Linguistics}, pages 7871--7880, Online. Association for
  Computational Linguistics.

\bibitem[{Li et~al.(2023)Li, Leng, Yan, Shen, Wang, MI, Fei, Feng, Yan, Wang,
  Zhan, Jia, Wu, and Sun}]{li2023chatharuhirevivinganimecharacter}
Cheng Li, Ziang Leng, Chenxi Yan, Junyi Shen, Hao Wang, Weishi MI, Yaying Fei,
  Xiaoyang Feng, Song Yan, HaoSheng Wang, Linkang Zhan, Yaokai Jia, Pingyu Wu,
  and Haozhen Sun. 2023.
\newblock \href {https://arxiv.org/abs/2308.09597} {{ChatHaruhi}: Reviving
  anime character in reality via large language model}.
\newblock \emph{Preprint}, arXiv:2308.09597.

\bibitem[{Li et~al.(2019)Li, Weston, and
  Roller}]{li2019acuteevalimproveddialogueevaluation}
Margaret Li, Jason Weston, and Stephen Roller. 2019.
\newblock \href {https://arxiv.org/abs/1909.03087} {{ACUTE-EVAL}: Improved
  dialogue evaluation with optimized questions and multi-turn comparisons}.
\newblock \emph{Preprint}, arXiv:1909.03087.

\bibitem[{Liao et~al.(2025)Liao, Ouyang, Lee, Yu, Tsai, and
  Yin}]{liao2025beyond}
Zikai Liao, Yi~Ouyang, Yi-Lun Lee, Chen-Ping Yu, Yi-Hsuan Tsai, and Zhaozheng
  Yin. 2025.
\newblock \href {https://arxiv.org/abs/2505.14654} {Beyond words: Multimodal
  {LLM} knows when to speak}.
\newblock \emph{Preprint}, arXiv:2505.14654.

\bibitem[{Lin(2004)}]{lin-2004-rouge}
Chin-Yew Lin. 2004.
\newblock \href {https://aclanthology.org/W04-1013/} {{ROUGE}: A package for
  automatic evaluation of summaries}.
\newblock In \emph{Text Summarization Branches Out}, pages 74--81, Barcelona,
  Spain. Association for Computational Linguistics.

\bibitem[{Liu et~al.(2021)Liu, Zheng, Demasi, Sabour, Li, Yu, Jiang, and
  Huang}]{liu-etal-2021-towards}
Siyang Liu, Chujie Zheng, Orianna Demasi, Sahand Sabour, Yu~Li, Zhou Yu, Yong
  Jiang, and Minlie Huang. 2021.
\newblock \href {https://doi.org/10.18653/v1/2021.acl-long.269} {Towards
  emotional support dialog systems}.
\newblock In \emph{Proceedings of the 59th Annual Meeting of the Association
  for Computational Linguistics and the 11th International Joint Conference on
  Natural Language Processing (Volume 1: Long Papers)}, pages 3469--3483,
  Online. Association for Computational Linguistics.

\bibitem[{Liu et~al.(2023)Liu, Iter, Xu, Wang, Xu, and
  Zhu}]{liu-etal-2023-geval}
Yang Liu, Dan Iter, Yichong Xu, Shuohang Wang, Ruochen Xu, and Chenguang Zhu.
  2023.
\newblock \href {https://doi.org/10.18653/v1/2023.emnlp-main.153} {{G-Eval}:
  {NLG} evaluation using {GPT-4} with better human alignment}.
\newblock In \emph{Proceedings of the 2023 Conference on Empirical Methods in
  Natural Language Processing}, pages 2511--2522, Singapore. Association for
  Computational Linguistics.

\bibitem[{Loshchilov and Hutter(2019)}]{loshchilov2018decoupled}
Ilya Loshchilov and Frank Hutter. 2019.
\newblock \href {https://openreview.net/forum?id=Bkg6RiCqY7} {Decoupled weight
  decay regularization}.
\newblock In \emph{International Conference on Learning Representations}.

\bibitem[{Maharana et~al.(2024)Maharana, Lee, Tulyakov, Bansal, Barbieri, and
  Fang}]{maharana-etal-2024-evaluating}
Adyasha Maharana, Dong-Ho Lee, Sergey Tulyakov, Mohit Bansal, Francesco
  Barbieri, and Yuwei Fang. 2024.
\newblock \href {https://doi.org/10.18653/v1/2024.acl-long.747} {Evaluating
  very long-term conversational memory of {LLM} agents}.
\newblock In \emph{Proceedings of the 62nd Annual Meeting of the Association
  for Computational Linguistics (Volume 1: Long Papers)}, pages 13851--13870,
  Bangkok, Thailand. Association for Computational Linguistics.

\bibitem[{Mehri et~al.(2022)Mehri, Choi, D'Haro, Deriu, Eskenazi, Gasic,
  Georgila, Hakkani-Tur, Li, Rieser, Shaikh, Traum, Yeh, Yu, Zhang, and
  Zhang}]{mehri2022reportnsffuturedirections}
Shikib Mehri, Jinho Choi, Luis~Fernando D'Haro, Jan Deriu, Maxine Eskenazi,
  Milica Gasic, Kallirroi Georgila, Dilek Hakkani-Tur, Zekang Li, Verena
  Rieser, Samira Shaikh, David Traum, Yi-Ting Yeh, Zhou Yu, Yizhe Zhang, and
  Chen Zhang. 2022.
\newblock \href {https://arxiv.org/abs/2203.10012} {Report from the {NSF}
  future directions workshop on automatic evaluation of dialog: Research
  directions and challenges}.
\newblock \emph{Preprint}, arXiv:2203.10012.

\bibitem[{Micikevicius et~al.(2018)Micikevicius, Narang, Alben, Diamos, Elsen,
  Garcia, Ginsburg, Houston, Kuchaiev, Venkatesh, and
  Wu}]{micikevicius2018mixed}
Paulius Micikevicius, Sharan Narang, Jonah Alben, Gregory Diamos, Erich Elsen,
  David Garcia, Boris Ginsburg, Michael Houston, Oleksii Kuchaiev, Ganesh
  Venkatesh, and Hao Wu. 2018.
\newblock \href {https://openreview.net/forum?id=r1gs9JgRZ} {Mixed precision
  training}.
\newblock In \emph{International Conference on Learning Representations}.

\bibitem[{Panickssery et~al.(2024)Panickssery, Bowman, and
  Feng}]{panickssery2024llm}
Arjun Panickssery, Samuel~R. Bowman, and Shi Feng. 2024.
\newblock \href {https://dl.acm.org/doi/10.5555/3737916.3740113} {{LLM}
  evaluators recognize and favor their own generations}.
\newblock In \emph{Proceedings of the 38th International Conference on Neural
  Information Processing Systems}, NIPS '24, Red Hook, NY, USA. Curran
  Associates Inc.

\bibitem[{Papineni et~al.(2002)Papineni, Roukos, Ward, and
  Zhu}]{papineni-etal-2002-bleu}
Kishore Papineni, Salim Roukos, Todd Ward, and Wei-Jing Zhu. 2002.
\newblock \href {https://doi.org/10.3115/1073083.1073135} {{B}leu: a method for
  automatic evaluation of machine translation}.
\newblock In \emph{Proceedings of the 40th Annual Meeting of the Association
  for Computational Linguistics}, pages 311--318, Philadelphia, Pennsylvania,
  USA. Association for Computational Linguistics.

\bibitem[{Qin et~al.(2021)Qin, Gupta, Upadhyay, He, Choi, and
  Faruqui}]{qin-etal-2021-timedial}
Lianhui Qin, Aditya Gupta, Shyam Upadhyay, Luheng He, Yejin Choi, and Manaal
  Faruqui. 2021.
\newblock \href {https://doi.org/10.18653/v1/2021.acl-long.549} {{TIMEDIAL}:
  Temporal commonsense reasoning in dialog}.
\newblock In \emph{Proceedings of the 59th Annual Meeting of the Association
  for Computational Linguistics and the 11th International Joint Conference on
  Natural Language Processing (Volume 1: Long Papers)}, pages 7066--7076,
  Online. Association for Computational Linguistics.

\bibitem[{Rashkin et~al.(2019)Rashkin, Smith, Li, and
  Boureau}]{rashkin-etal-2019-towards}
Hannah Rashkin, Eric~Michael Smith, Margaret Li, and Y-Lan Boureau. 2019.
\newblock \href {https://doi.org/10.18653/v1/P19-1534} {Towards empathetic
  open-domain conversation models: A new benchmark and dataset}.
\newblock In \emph{Proceedings of the 57th Annual Meeting of the Association
  for Computational Linguistics}, pages 5370--5381, Florence, Italy.
  Association for Computational Linguistics.

\bibitem[{Roller et~al.(2021)Roller, Dinan, Goyal, Ju, Williamson, Liu, Xu,
  Ott, Smith, Boureau, and Weston}]{roller-etal-2021-recipes}
Stephen Roller, Emily Dinan, Naman Goyal, Da~Ju, Mary Williamson, Yinhan Liu,
  Jing Xu, Myle Ott, Eric~Michael Smith, Y-Lan Boureau, and Jason Weston. 2021.
\newblock \href {https://doi.org/10.18653/v1/2021.eacl-main.24} {Recipes for
  building an open-domain chatbot}.
\newblock In \emph{Proceedings of the 16th Conference of the European Chapter
  of the Association for Computational Linguistics: Main Volume}, pages
  300--325, Online. Association for Computational Linguistics.

\bibitem[{Shao et~al.(2023)Shao, Li, Dai, and Qiu}]{shao-etal-2023-character}
Yunfan Shao, Linyang Li, Junqi Dai, and Xipeng Qiu. 2023.
\newblock \href {https://aclanthology.org/2023.emnlp-main.814/}
  {Character-{LLM}: A trainable agent for role-playing}.
\newblock In \emph{Proceedings of the 2023 Conference on Empirical Methods in
  Natural Language Processing}, pages 13153--13187, Singapore. Association for
  Computational Linguistics.

\bibitem[{Tsunomori et~al.(2023)Tsunomori, Ishihata, and
  Sugiyama}]{tsunomori-etal-2023-time}
Yuiko Tsunomori, Masakazu Ishihata, and Hiroaki Sugiyama. 2023.
\newblock \href {https://doi.org/10.18653/v1/2023.findings-emnlp.341}
  {Time-considerable dialogue models via reranking by time dependency}.
\newblock In \emph{Findings of the Association for Computational Linguistics:
  EMNLP 2023}, pages 5136--5149, Singapore. Association for Computational
  Linguistics.

\bibitem[{Wang et~al.(2024)Wang, Cheng, Song, Zhang, and
  Niu}]{niu-etal-2024-enhancing}
Xingguang Wang, Xuxin Cheng, Juntong Song, Tong Zhang, and Cheng Niu. 2024.
\newblock \href {https://doi.org/10.18653/v1/2024.acl-long.473} {Enhancing
  dialogue state tracking models through {LLM}-backed user-agents simulation}.
\newblock In \emph{Proceedings of the 62nd Annual Meeting of the Association
  for Computational Linguistics (Volume 1: Long Papers)}, pages 8724--8741,
  Bangkok, Thailand. Association for Computational Linguistics.

\bibitem[{Wang et~al.(2025)Wang, Meng, Wang, Zhang, and
  Zhao}]{wang2025proactivevideoqa}
Yueqian Wang, Xiaojun Meng, Yifan Wang, Huishuai Zhang, and Dongyan Zhao. 2025.
\newblock \href {https://arxiv.org/abs/2507.09313} {{ProactiveVideoQA}: A
  comprehensive benchmark evaluating proactive interactions in video large
  language models}.
\newblock \emph{Preprint}, arXiv:2507.09313.

\bibitem[{Wataoka et~al.(2024)Wataoka, Takahashi, and Ri}]{wataoka2024self}
Koki Wataoka, Tsubasa Takahashi, and Ryokan Ri. 2024.
\newblock \href {https://openreview.net/forum?id=tLZZZIgPJX} {Self-preference
  bias in {LLM}-as-a-judge}.
\newblock In \emph{{NeurIPS} Safe Generative AI Workshop 2024}.

\bibitem[{Wei et~al.(2022)Wei, Wang, Schuurmans, Bosma, Ichter, Xia, Chi, Le,
  and Zhou}]{10.5555/3600270.3602070}
Jason Wei, Xuezhi Wang, Dale Schuurmans, Maarten Bosma, Brian Ichter, Fei Xia,
  Ed~H. Chi, Quoc~V. Le, and Denny Zhou. 2022.
\newblock \href {https://dl.acm.org/doi/10.5555/3600270.3602070}
  {Chain-of-thought prompting elicits reasoning in large language models}.
\newblock In \emph{Proceedings of the 36th International Conference on Neural
  Information Processing Systems}, NIPS '22, Red Hook, NY, USA. Curran
  Associates Inc.

\bibitem[{Xu et~al.(2022{\natexlab{a}})Xu, Szlam, and
  Weston}]{xu-etal-2022-beyond}
Jing Xu, Arthur Szlam, and Jason Weston. 2022{\natexlab{a}}.
\newblock \href {https://doi.org/10.18653/v1/2022.acl-long.356} {Beyond
  goldfish memory: Long-term open-domain conversation}.
\newblock In \emph{Proceedings of the 60th Annual Meeting of the Association
  for Computational Linguistics (Volume 1: Long Papers)}, pages 5180--5197,
  Dublin, Ireland. Association for Computational Linguistics.

\bibitem[{Xu et~al.(2022{\natexlab{b}})Xu, Gou, Wu, Niu, Wu, Wang, and
  Wang}]{xu-etal-2022-long-cr}
Xinchao Xu, Zhibin Gou, Wenquan Wu, Zheng-Yu Niu, Hua Wu, Haifeng Wang, and
  Shihang Wang. 2022{\natexlab{b}}.
\newblock \href {https://doi.org/10.18653/v1/2022.findings-acl.207} {Long time
  no see! {O}pen-domain conversation with long-term persona memory}.
\newblock In \emph{Findings of the Association for Computational Linguistics:
  ACL 2022}, pages 2639--2650, Dublin, Ireland. Association for Computational
  Linguistics.

\bibitem[{Yang et~al.(2025{\natexlab{a}})Yang, Xu, Zeng, Liu, Jiang, Lu, Chen,
  Jiang, Xing, and Yan}]{yang2025contextagent}
Bufang Yang, Lilin Xu, Liekang Zeng, Kaiwei Liu, Siyang Jiang, Wenrui Lu,
  Hongkai Chen, Xiaofan Jiang, Guoliang Xing, and Zhenyu Yan.
  2025{\natexlab{a}}.
\newblock \href {https://openreview.net/forum?id=tRXt10xKc5} {{ContextAgent}:
  Context-aware proactive {LLM} agents with open-world sensory perceptions}.
\newblock In \emph{The Thirty-ninth Annual Conference on Neural Information
  Processing Systems}.

\bibitem[{Yang et~al.(2025{\natexlab{b}})Yang, Li, Fang, and
  Chen}]{yang-etal-2025-mtpchat}
Wanqi Yang, Yanda Li, Meng Fang, and Ling Chen. 2025{\natexlab{b}}.
\newblock \href {https://doi.org/10.18653/v1/2025.findings-naacl.323}
  {{MTPC}hat: A multimodal time-aware persona dataset for conversational
  agents}.
\newblock In \emph{Findings of the Association for Computational Linguistics:
  NAACL 2025}, pages 5815--5826, Albuquerque, New Mexico. Association for
  Computational Linguistics.

\bibitem[{Yoon et~al.(2024)Yoon, He, Echterhoff, and
  McAuley}]{yoon-etal-2024-evaluating}
Se-eun Yoon, Zhankui He, Jessica Echterhoff, and Julian McAuley. 2024.
\newblock \href {https://doi.org/10.18653/v1/2024.naacl-long.83} {Evaluating
  large language models as generative user simulators for conversational
  recommendation}.
\newblock In \emph{Proceedings of the 2024 Conference of the North American
  Chapter of the Association for Computational Linguistics: Human Language
  Technologies (Volume 1: Long Papers)}, pages 1490--1504, Mexico City, Mexico.
  Association for Computational Linguistics.

\bibitem[{Yoon et~al.(2025)Yoon, Son, So, Kim, Cho, Park, Lee, and
  Kim}]{yoon-etal-2025-beyond}
Yejin Yoon, Yuri Son, Namyoung So, Minseo Kim, Minsoo Cho, Chanhee Park,
  Seungshin Lee, and Taeuk Kim. 2025.
\newblock \href {https://doi.org/10.18653/v1/2025.emnlp-main.672} {Beyond
  task-oriented and chitchat dialogues: Proactive and transition-aware
  conversational agents}.
\newblock In \emph{Proceedings of the 2025 Conference on Empirical Methods in
  Natural Language Processing}, pages 13291--13317, Suzhou, China. Association
  for Computational Linguistics.

\bibitem[{Zhang et~al.(2023)Zhang, Naradowsky, and
  Miyao}]{zhang-etal-2023-mind}
Qiang Zhang, Jason Naradowsky, and Yusuke Miyao. 2023.
\newblock \href {https://doi.org/10.18653/v1/2023.findings-emnlp.720} {Mind the
  gap between conversations for improved long-term dialogue generation}.
\newblock In \emph{Findings of the Association for Computational Linguistics:
  EMNLP 2023}, pages 10735--10762, Singapore. Association for Computational
  Linguistics.

\bibitem[{Zhang et~al.(2018)Zhang, Dinan, Urbanek, Szlam, Kiela, and
  Weston}]{zhang-etal-2018-personalizing}
Saizheng Zhang, Emily Dinan, Jack Urbanek, Arthur Szlam, Douwe Kiela, and Jason
  Weston. 2018.
\newblock \href {https://doi.org/10.18653/v1/P18-1205} {Personalizing dialogue
  agents: {I} have a dog, do you have pets too?}
\newblock In \emph{Proceedings of the 56th Annual Meeting of the Association
  for Computational Linguistics (Volume 1: Long Papers)}, pages 2204--2213,
  Melbourne, Australia. Association for Computational Linguistics.

\bibitem[{Zhang et~al.(2024)Zhang, Zhang, Zhao, Zhou, and
  Jin}]{zhang-etal-2024-escot}
Tenggan Zhang, Xinjie Zhang, Jinming Zhao, Li~Zhou, and Qin Jin. 2024.
\newblock \href {https://doi.org/10.18653/v1/2024.acl-long.723} {{ESC}o{T}:
  Towards interpretable emotional support dialogue systems}.
\newblock In \emph{Proceedings of the 62nd Annual Meeting of the Association
  for Computational Linguistics (Volume 1: Long Papers)}, pages 13395--13412,
  Bangkok, Thailand. Association for Computational Linguistics.

\bibitem[{Zhang et~al.(2020)Zhang, Kishore, Wu, Weinberger, and
  Artzi}]{DBLP:conf/iclr/ZhangKWWA20}
Tianyi Zhang, Varsha Kishore, Felix Wu, Kilian~Q. Weinberger, and Yoav Artzi.
  2020.
\newblock \href {https://openreview.net/forum?id=SkeHuCVFDr} {{BERTScore}:
  Evaluating text generation with {BERT}}.
\newblock In \emph{8th International Conference on Learning Representations,
  {ICLR} 2020, Addis Ababa, Ethiopia, April 26-30, 2020}. OpenReview.net.

\bibitem[{Zhou et~al.(2019)Zhou, Khashabi, Ning, and
  Roth}]{zhou-etal-2019-going-cr}
Ben Zhou, Daniel Khashabi, Qiang Ning, and Dan Roth. 2019.
\newblock \href {https://doi.org/10.18653/v1/D19-1332} {``{G}oing on a
  vacation'' takes longer than ``{G}oing for a walk'': A study of temporal
  commonsense understanding}.
\newblock In \emph{Proceedings of the 2019 Conference on Empirical Methods in
  Natural Language Processing and the 9th International Joint Conference on
  Natural Language Processing (EMNLP-IJCNLP)}, pages 3363--3369, Hong Kong,
  China. Association for Computational Linguistics.

\bibitem[{Zhou et~al.(2021)Zhou, Gopalakrishnan, Hedayatnia, Kim, Pujara, Ren,
  Liu, and Hakkani-Tur}]{zhou-etal-2021-commonsense}
Pei Zhou, Karthik Gopalakrishnan, Behnam Hedayatnia, Seokhwan Kim, Jay Pujara,
  Xiang Ren, Yang Liu, and Dilek Hakkani-Tur. 2021.
\newblock \href {https://doi.org/10.18653/v1/2021.sigdial-1.13}
  {Commonsense-focused dialogues for response generation: An empirical study}.
\newblock In \emph{Proceedings of the 22nd Annual Meeting of the Special
  Interest Group on Discourse and Dialogue}, pages 121--132, Singapore and
  Online. Association for Computational Linguistics.

\bibitem[{Zhou et~al.(2024)Zhou, Zhu, Mathur, Zhang, Yu, Qi, Morency, Bisk,
  Fried, Neubig, and Sap}]{zhou2024sotopia}
Xuhui Zhou, Hao Zhu, Leena Mathur, Ruohong Zhang, Haofei Yu, Zhengyang Qi,
  Louis-Philippe Morency, Yonatan Bisk, Daniel Fried, Graham Neubig, and
  Maarten Sap. 2024.
\newblock \href {https://openreview.net/forum?id=mM7VurbA4r} {{SOTOPIA}:
  Interactive evaluation for social intelligence in language agents}.
\newblock In \emph{The Twelfth International Conference on Learning
  Representations}.

\end{thebibliography}

\clearpage
\appendix

\section{Prevalence of Delayed Responses in Real Conversations}
\label{appx:delay_prevalence}

To motivate the \textit{when-to-respond} problem with real human behavior, we measure the distribution of the reply time gaps in large-scale, publicly available conversation corpora.
We aggregate three real messaging corpora (a social messaging corpus, an online conversation corpus, and a messenger-style corpus) from AI Hub\footnote{\url{https://aihub.or.kr/}}, comprising 12.7M reply turns in total.
Note that these corpora are in a different language (Korean) from our benchmark.
We use them as cross-lingual evidence that delayed responding is a general phenomenon in human messaging.

\begin{figure}[t]
\centering
\begin{subfigure}{0.49\columnwidth}
  \centering
  \includegraphics[width=\linewidth]{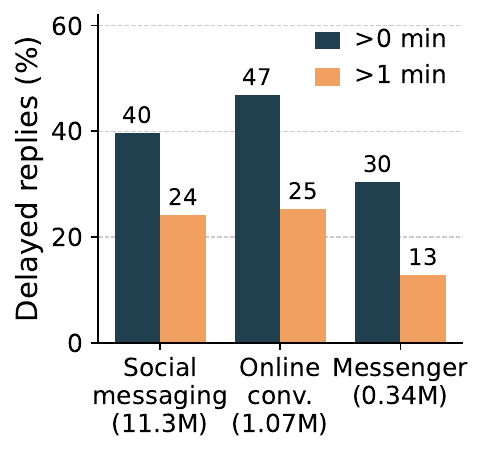}
  \caption{Delayed replies (gap $>\!0$ and $>\!1$ min) per corpus.}
  \label{fig:delay_prevalence_a}
\end{subfigure}
\hfill
\begin{subfigure}{0.49\columnwidth}
  \centering
  \includegraphics[width=\linewidth]{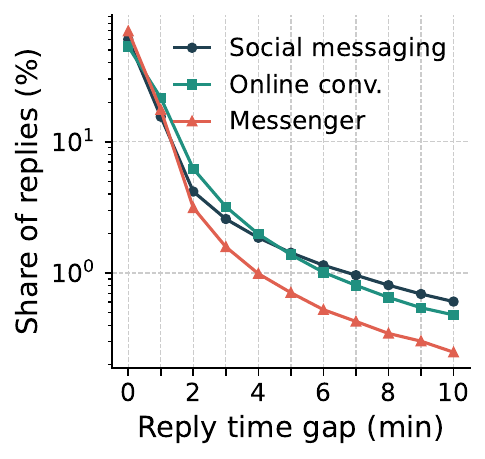}
  \caption{Reply gap distribution over $0$--$10$ min (log scale).}
  \label{fig:delay_prevalence_b}
\end{subfigure}
\caption{Reply time gap statistics in real messaging corpora. Most replies are instant ($0$ min), yet a large minority are delayed, and the gap distribution has a long tail extending well beyond the instant bin.}
\label{fig:delay_prevalence}
\end{figure}

As summarized in Figure~\ref{fig:delay_prevalence}, while a large fraction of replies are effectively instant (gap $=0$~min), roughly 40\% of replies are delayed (gap $>0$~min), and about 24\% remain delayed under a conservative threshold that treats gaps of up to one minute as instant.
A separate large-scale messaging log of 25.6M time differences shows a consistent pattern, with 27\% of replies delayed.
These statistics establish that delayed responding is pervasive in real conversation. An agent that can only respond instantly is therefore a poor default model of human messaging behavior, which motivates modeling \textit{when to respond}.

We emphasize what this analysis does and does not show.
It establishes a necessary condition that response delays are common and thus worth modeling, but it does not attribute these delays to any single cause.
Delays in raw corpora arise from many factors, including user unavailability, platform asynchrony, social norms, and the event-grounded situations that are the focus of this work.
Since the event-grounded subset cannot be reliably isolated from raw logs by any mechanical filter, we instead synthesize it under controlled constraints that ensure a delay is warranted by the ongoing event.

\section{Data Construction Details}
\label{appx:data_construction_details}
\subsection{Data Construction Prompts}
\label{appx:chatgpt_prompts}

We provide prompts used for data construction processes of both evaluation and training datasets.
The contents within curly brackets represent the corresponding elements for each example.
The statements used for ATOMIC$_{20}^{20}$ duration estimation are constructed by concatenating the head and tail with a conjunction that represents each relation category.
We present the prompts in the order of the processes, where the output of each step serves as the input for the next step.
The dialogue generation prompt asks for the delayed response before the instant one, and Appendix~\ref{appx:order_bias} finds no significant effect of this order on the generated candidates.

\begin{tcolorbox}[colback=green!5,colframe=green!40!black,title=ATOMIC$_{20}^{20}$ Duration Estimation,enhanced,breakable]
You are given a statement about common events in our daily lives. Your task is to estimate the typical duration of the key event in the form of (quantity of time + unit) (e.g., seconds, minutes, hours, days, weeks, months, years, decades, or centuries) based on the temporal common sense of average humans. \\
\\
\lbrack Examples\rbrack \\
Statement: After dinner, he went to look for Max one last time before he had to take a bath and go to bed. \\
Key event: having dinner \\
Duration: 1 hour \\
\\
Statement: Jennie and Bryan boarded a 6:00 A.M. flight from Seoul to Los Angeles International Airport. \\
Key event: flight from Seoul to Los Angeles \\
Duration: 12 hours \\
\\
Event: Carl Laemmle, head of Universal Studios, gave Einstein a tour of his studio and introduced him to Chaplin. \\
Key event: tour of his studio \\
Duration: 45 minutes \\
\lbrack End of Examples\rbrack \\
\\
Statement: \{statement\}
\end{tcolorbox}

\begin{tcolorbox}[colback=green!5,colframe=green!40!black,title=MC-TACO Event Descriptions,enhanced,breakable]
You are given an event and a question and answer for the duration that denotes how much time is needed for the event to happen. \\
Write a story regarding the event in one sentence. \\
\\
Sentence: \{sentence\} \\
Question: \{question\} \\
Answer: \{duration\} \\
\end{tcolorbox}

\begin{tcolorbox}[colback=green!5,colframe=green!40!black,title=ATOMIC$_{20}^{20}$ Event Descriptions,enhanced,breakable]
You are given a statement, the key event and the duration that denotes how much time is needed for the event to happen. \\
Write a story regarding the event in one sentence. \\
\\
Statement: \{statement\} \\
Key event: \{event\} \\
Duration: \{duration\}
\end{tcolorbox}

\begin{tcolorbox}[colback=green!5,colframe=green!40!black,title=Dialogue Generation,enhanced,breakable]
You are given an event narrative and the duration. Your task is to create an instant message dialog between two speakers. The following conditions MUST be met. \\
\\
\lbrack Instructions\rbrack \\
1. Speaker \{A,B\} is in the middle of the event now, while speaker \{B,A\} is physically apart from. \\
2. Do not directly mention the duration in the dialog. \\
3. After \{B,A\}'s last turn, add "\lbrack \{duration\} later\rbrack", where duration is the amount of time passed in real world. \\
4-1. Generate \{A,B\}'s last message which is timely as if \{A,B\} spent time to finish the event. \\
4-2. In contrast, generate \{A,B\}'s last message as if \{A,B\} is responding instantaneously right before the event to happen. \\
Make sure that the timely response and the instantaneous response are time-situationally different. \\
\lbrack End of Instructions\rbrack \\
\\
\lbrack Example\rbrack \\
\{dialog example\} \\
\lbrack End of Example\rbrack \\
\\
Narrative: \{event description\} \\
Duration: \{duration\}
\end{tcolorbox}

\subsection{Few-shot Examples}
\label{appx:few-shot_examples}
We provide six author-written dialogue examples randomly fed into GPT-4o as one-shot demonstrations when generating dialogues for \textsc{TimelyChat} using the MC-TACO dataset.

\begin{tcolorbox}[colback=green!5,colframe=green!40!black,title=5-turn Dialogue,enhanced,breakable]
Narrative: After dinner, he took a shower before he went to bed.\\
Duration: 20 minutes \\
\\
A: I finally got home. What a day! \\
B: It's eleven p.m. and you just got back home? It must be very tough day today. \\
A: Whooa Imma take a shower. I'm too tired. \\
B: Wash out all your fatigue with it. \\
\lbrack 20 minutes later\rbrack \\
(delayed response) \\
A: I feel much better now! You didn't go to bed? \\
(instantaneous response) \\
A: How nice of you :) Give me a moment. brb
\end{tcolorbox}

\begin{tcolorbox}[colback=green!5,colframe=green!40!black,title=6-turn Dialogue,enhanced,breakable]
Narrative: She has taken calculus class and she had a final exam. \\
Duration: 2 hours \\
\\
A: Hey, what are you up to? \\
B: I'm gonna take the calculus final exam in 20 minutes. I feel so nervous. \\
A: You studied really hard, didn't you? I'm 100\% sure you'll do well. \\
B: But the last two chapters were too difficult for me to understand. \\
A: That means others feel the same. Don't worry too much! \\
\lbrack 2 hours later\rbrack \\
(delayed response) \\
B: It wasn't much harder than I expected. I hope I get a good grade. \\
(instantaneous response) \\
B: Thank you for cheering me up. I hope the exam is not that hard.
\end{tcolorbox}

\begin{tcolorbox}[colback=green!5,colframe=green!40!black,title=7-turn Dialogue,enhanced,breakable]
Narrative: He enjoyed working out at the gym. \\
Duration: 1 hour 30 minutes \\
\\
A: I'm going to the gym now. Wanna join me? \\
B: I don't feel like working out today. Sorry. \\
A: You don't feel good? What happened? \\
B: I played football so hard yesterday that I can't even walk right. \\
A: Okay, I understand. Maybe next time! \\
B: Enjoy your routine! I think I can make it tomorrow. \\
\lbrack 2 hours later\rbrack \\
(delayed response) \\
A: I focused on my lower body today. Chest tomorrow? \\
(instantaneous response) \\
A: Gonna work out hard on my lower body. Chest tomorrow?
\end{tcolorbox}

\begin{tcolorbox}[colback=green!5,colframe=green!40!black,title=8-turn Dialogue,enhanced,breakable]
Narrative: She had felt so tired that she went to bed right after the tv show. \\
Duration: 8 hours \\
\\
A: Are you watching the saturday night live? \\
B: I'm watching it now but I'm too tired. \\
A: I didn't expect today's host is such a comedian lol \\
B: Yeah almost the end of the show. I feel like going to bed little bit early. \\
A: What made you so tired? You had any plan? \\
B: I went to an amusement park with my sister. We had a really good time there. \\
A: Oh I see. Think I should let you go. Sleep tight! \\
\lbrack 8 hours later\rbrack \\
(delayed response) \\
B: Good morning. Did you sleep tight, too? \\
(instantaneous response) \\
B: Good night. I'll text you in the morning.
\end{tcolorbox}

\begin{tcolorbox}[colback=green!5,colframe=green!40!black,title=9-turn Dialogue,enhanced,breakable]
Narrative: He took an intercity bus to get to his hometown. \\
Duration: 5 hours \\
\\
A: What are you going to do on these holidays? \\
B: My parents and I usually have dinner together on the Eve. \\
A: Me too. So I'm heading to my town right now. \\
B: How do you get there? By bus or train? \\
A: I used to take trains, but I take an intercity bus for this time. \\
B: Why? the tickets' been already sold out? \\
A: Unfortunately yes... It will take little bit longer. \\
B: Have a nice trip though. Your family must be waiting for you. \\
\lbrack 5 hours later\rbrack \\
(delayed response) \\
A: Finally I'm back at home! It took almost 5 hours. \\
(instantaneous response) \\
A: I'm gonna sleep all along in the bus. See you a few hours later.
\end{tcolorbox}

\begin{tcolorbox}[colback=green!5,colframe=green!40!black,title=10-turn Dialogue,enhanced,breakable]
Narrative: She played the popular online game with her friends. \\
Duration: 30 minutes \\
\\
A: Have you heard of the League of Legends? \\
B: Absolutely! I play it almost everyday with my classmates. \\
A: I've heard of, but I've never played if before. \\
B: We have a game soon. Wanna join us? \\
A: Isn't it a team game? I'm not a good gamer though. \\
B: It's not a big deal. They will welcome you. \\
A: Well, maybe next time. I need to play it by myself first. \\
B: How about getting tutorial with me after this? I'll teach you. \\
A: Sounds good. Enjoy your game with your teammates. \\
\lbrack 30 minutes later\rbrack \\
(delayed response) \\
B: We won! The game was nip and tuck. We were so close to losing. \\
(instantaneous response) \\
B: I'll be back just after the game. Wish me a good luck!
\end{tcolorbox}

\subsection{Time Interval Distribution}
\label{appx:interval_dist}

We semantically categorize the durations and present the distributions of time intervals in the two proposed datasets in Table~\ref{tab:interval_distribution} for a comprehensive comparison.

\begin{table}[ht]
\centering
\begin{tabular}{@{}lcc@{}}
\toprule
Time Interval           & Train  & Test   \\ \midrule
0--5 min (instant)       & 18.5\% & 5.6\%  \\
5--30 min (short)        & 42.3\% & 38.6\% \\
30 min--2 hrs (moderate) & 23.8\% & 32.1\% \\
2--6 hrs (long)          & 2.9\%  & 13.9\% \\
6--12 hrs (half-day)     & 2.0\%  & 5.2\%  \\
12--24 hrs (full-day)    & 10.5\% & 4.6\%  \\ \bottomrule
\end{tabular}%
\caption{Time interval distribution of the augmented training dataset (Train) and the \textsc{TimelyChat} benchmark (Test).}
\label{tab:interval_distribution}
\end{table}

A predicted interval is a point estimate of a duration that can vary even for the same event, so we measure how close each prediction falls to the ground truth on the six ranges of Table~\ref{tab:interval_distribution}.
On the \textsc{TimelyChat} benchmark, \textsc{Timer} 3B places 38.0\% of its predictions in the correct range and 77.8\% within a one-range difference, indicating that most of its errors are near-misses on an ordinal scale.

\subsection{Training Set Quality Assessment}
\label{appx:trainset_audit}

The 55K training set is generated without manual screening, so we measure its quality.
Two of the authors independently judge 200 dialogues drawn from the validation split.
We check each dialogue against the three generation conditions of Section~\ref{sec:data_construction} and two additional conditions, \emph{speaker consistency} (each turn is attributed to the speaker the scenario assigns it to) and \emph{duration validity} (the pseudo-labeled duration is plausible for the event and falls within the one-minute to 24-hour range).

\begin{table}[t]
\centering
\begin{tabular}{@{}lc@{}}
\toprule
Condition & Pass rate \\ \midrule
Spatial separation     & 98.0\% \\
Temporal implicitness  & 92.8\% \\
Mutual exclusivity     & 94.8\% \\
Speaker consistency    & 100.0\% \\
Duration validity      & 95.5\% \\ \bottomrule
\end{tabular}
\caption{Per-condition pass rates from the training set assessment (two annotators, $n=200$). Raw inter-annotator agreement is 99.4\% and Cohen's $\kappa$ is 0.918.}
\label{tab:trainset_audit}
\end{table}

Table~\ref{tab:trainset_audit} reports the per-condition pass rates.
Raw agreement between the two annotators is 99.4\% and Cohen's $\kappa$ is 0.918.
Temporal implicitness is the weakest condition at 92.8\%.
This is the expected failure mode, since it is the only condition that restricts what the response may \emph{not} say and the generator occasionally mentions the elapsed time verbatim.

\subsection{Presentation Order in Dialogue Generation}
\label{appx:order_bias}

Since the dialogue generation prompt asks for the delayed candidate before the instant one, we test the effect of this order by reversing it.
For 50 narratives, we regenerate both final-turn candidates under both instruction orders.
We compare the candidates in two aspects, their length in words and their time leakage, i.e., whether they mention an explicit time span such as ``10 minutes''.
We use GPT-4o and Claude Sonnet 4.5, and compare the two orders with Welch's $t$-test.

\begin{table}[t]
\centering
\resizebox{\columnwidth}{!}{%
\begin{tabular}{@{}llcccccc@{}}
\toprule
 & & \multicolumn{3}{c}{Length (words)} & \multicolumn{3}{c}{Leakage (\%)} \\
\cmidrule(lr){3-5}\cmidrule(lr){6-8}
Model & Cand. & D-first & I-first & $p$ & D-first & I-first & $p$ \\ \midrule
\multirow{2}{*}{GPT-4o} & Del. & 32.0 & 30.1 & 0.362 & 4 & 2 & 0.563 \\
                        & Inst. & 22.9 & 20.3 & 0.059 & 0 & 0 & ---   \\ \midrule
\multirow{2}{*}{Sonnet 4.5} & Del. & 24.7 & 26.2 & 0.308 & 4 & 2 & 0.563 \\
                        & Inst. & 18.9 & 18.7 & 0.915 & 0 & 2 & 0.322 \\ \bottomrule
\end{tabular}%
}
\caption{Mean length and time leakage rate of the regenerated candidates under the two instruction orders, with $p$-values of Welch's $t$-test between the orders ($n=50$). D-first and I-first denote the delayed-first and instant-first instruction orders. The test is undefined for the time leakage of the instant candidate under GPT-4o, which is 0\% under both orders.}
\label{tab:order_bias}
\end{table}

As shown in Table~\ref{tab:order_bias}, neither aspect differs significantly between the two orders under either model.
The smallest $p$-value is for the length of the instant candidate under GPT-4o ($p=0.059$), but it does not reproduce under Claude Sonnet 4.5 ($p=0.915$), so we regard it as sampling noise.
The time leakage rate also lies between 0\% and 4\% in all settings, which is consistent with the temporal implicitness condition.

\section{Evaluation Details}
\label{appx:evaluation_setup_details}
\subsection{Implementation Details}
\label{appx:implementation_details}

\paragraph{Instruction-tuned Models.}
We use the vLLM library\footnote{\url{https://docs.vllm.ai}} for the inference of LLaMA 3.1 Instruct 8B and 70B on 4 NVIDIA A100 80GB GPUs.
All instruction-tuned models employ top-$p$ sampling with temperature 1.0 and $p=0.95$ during inference.
We provide the prompts used for in-context learning methods on both response timing prediction and time-conditioned response generation below.

\begin{tcolorbox}[colback=green!5,colframe=green!40!black,title=Prompt for Response Timing Prediction,enhanced,breakable]
You are given a conversation between two speakers. \\
Your task is to estimate a time interval needed until the next response, considering the duration of the event in the conversation ranging from 0 minutes to 24 hours (1 day). \\
If the next response is expected to be immediate, you will output "0 minutes". \\
Otherwise, you will output a digit and a unit of time (e.g., 5 minutes, 2 hours). \\
Just output the time interval without any other text. \\
\\
\lbrack Example $n$\rbrack \\
\{few-shot OR CoT example\} \\
\\
\#\#\# Dialog context \#\#\# \\
\{context\} \\
\\
Answer format: n (0<=n<=1440) minutes \\
The estimated time interval is:
\end{tcolorbox}

\begin{tcolorbox}[colback=green!5,colframe=green!40!black,title=Prompt for Time-conditioned Response Generation,enhanced,breakable]
You are given a conversation between two speakers and the elapsed time since the last utterance. \\
Your task is to generate the next response that aligns well with the temporal context represented by the time interval in parentheses. \\
Just output the response without any other text. \\
\\
\lbrack Example $n$\rbrack \\
\{few-shot OR CoT example\} \\
\\
\#\#\# Dialog context \#\#\# \\
\{context\} \\
\\
\#\#\# Next response \#\#\# \\
\{target speaker\}: (\{time interval\} later)
\end{tcolorbox}

\paragraph{Time-aware Models.}
We use the Hugging Face library\footnote{\url{https://huggingface.co}} for the inference of MSC 3B and ReBot 400M\footnote{\url{https://huggingface.co/jihyoung/rebot-generation}}.
We convert MSC 3B on the ParlAI framework\footnote{\url{https://parl.ai}} into a Hugging Face checkpoint.
We initialize \textsc{Timer} 3B from COSMO 3B~\citep{kim-etal-2023-soda}, a T5-based dialogue model, and fine-tune GapChat 3B and \textsc{Timer} 3B on their respective training data using the DeepSpeed library\footnote{\url{https://www.deepspeed.ai}} with mixed precision training~\citep{micikevicius2018mixed}.
We train the models for 3 epochs using the AdamW optimizer~\citep{loshchilov2018decoupled} with a learning rate of $1\times10^{-4}$, running on 2 NVIDIA A100 80GB GPUs for 9 hours.
We use $\lambda=1.0$ as a balanced scale factor of the two losses when training \textsc{Timer} 3B.
During inference, we apply beam search with a beam size of 3 and top-$p$ sampling with $p=0.95$.

\subsection{User Simulator Prompts}
\label{appx:user_simulator_prompts}
We present the prompt fed into GPT-4o to create the user simulator.

\begin{tcolorbox}[colback=green!5,colframe=green!40!black,title=User Simulator Prompt,enhanced,breakable]
You are a user simulator (user) engaging in an event-driven dialog with a dialog agent (agent). \\
Given the dialog context, your task is to proceed the conversation by one turn under the following assumptions: \\
1. agent responds after the elapsed time specified in the parentheses from the previous user utterance. If the delay is "0 minutes", agent is assumed to respond immediately. \\
2. user is assumed to respond to agent without any delay. \\
\\
Conversation: \\
\{context\}
\end{tcolorbox}

\subsection{LLM-as-a-Judge Details}
\label{appx:evaluation_criteria}

We elucidate the LLM-as-a-judge prompts used in turn-level and dialogue-level evaluations, along with the evaluation criteria and steps for each metric.

\begin{tcolorbox}[colback=green!5,colframe=green!40!black,title=Turn-level Prompt,enhanced,breakable]
You will be given a conversation between two individuals via messaging, along with the elapsed time since the last utterance. You will then be given a potential response for the next turn. \\
Your task is to rate the response on one metric. Please make sure you read and understand these instructions carefully. \\
\\
Evaluation Criteria: \\
\{metric\} (1-5): \{criteria\} \\
\\
Evaluation Steps: \\
\{steps\}
\end{tcolorbox}

\begin{tcolorbox}[colback=green!5,colframe=green!40!black,title=Dialogue-level Prompt,enhanced,breakable]
You will be given a conversation between a dialog agent and a user. \\
Throughout the conversation, the agent proactively determines the delay of its response to the user's previous message, simulating delayed responses due to event experiences that take certain time to process. \\
At each agent's turn, the delay is provided in the parentheses followed by the message. \\
Your task is to rate the dialog agent on one metric. Please make sure you read and understand these instructions carefully. \\
\\
Evaluation Criteria: \\
\{metric\} (1-5): \{criteria\} \\
\\
Evaluation Steps: \\
\{steps\}
\end{tcolorbox}

\paragraph{Evaluation Criteria and Steps.}
\begin{itemize}[leftmargin=*,topsep=2pt,itemsep=2pt,parsep=0pt]
    \item \textbf{Naturalness} (1-5): the extent to which the response reads naturally given the dialogue context. \\1. Assess the flow and coherence of the response in the conversation: Consider how seamlessly the response connects with the previous message. \\2. Evaluate the tone and style compatibility: Determine if the response's tone and style match those of the previous messages. \\3. Rate on a scale from 1 to 5, where 1 indicates the response is unnatural or inappropriate, and 5 indicates a perfectly natural continuation of the conversation.
    \item (Turn-level) \textbf{Time-specificity} (1-5): the extent to which the response ONLY makes sense when the specified time has passed, contrary to a time-agnostic response that makes sense regardless of time. \\1. Read the provided conversation and take note of the elapsed time since the previous message. \\2. Consider the context of the conversation, focusing on how the passage of time might affect the relevance or appropriateness of the response. \\3. Evaluate whether the potential response provided is time-specific. That is, determine if the response directly relates to or is clearly influenced by the elapsed time between the last utterance and the response. \\4. Rate on a scale from 1 to 5, where 1 indicates the response is completely time-agnostic and unaffected by the passage of time, and 5 indicates the response is entirely time-specific; it only makes sense because of the amount of time that has passed since the previous message.
    \item \textbf{Coherence} (1-5): the extent to which the agent maintains a good conversation flow. \\1. Assess the flow and coherence of the agent's responses in the conversation. \\2. Evaluate the tone and style compatibility throughout the conversation. \\3. Rate on a scale from 1 to 5, where 1 indicates the agent's responses are incoherent or inappropriate, and 5 indicates the agent's responses are perfectly coherent and appropriate.
    \item \textbf{Delay-appropriateness} (1-5): the extent to which the agent poses delays with appropriate frequency and amount. \\1. Assess whether the agent poses unnecessary or excessively frequent delays that could harm the conversation flow. \\2. Evaluate whether the amounts of delays (if not 0 minutes) reflect the typical duration of events implied in the corresponding message. \\3. Rate on a scale from 1 to 5, where 1 indicates the agent overuses and misuses delays, and 5 indicates the agent uses delays appropriately in terms of frequency and amount.
    \item (Dialogue-level) \textbf{Time-specificity} (1-5): the extent to which the agent's responses ONLY make sense when the specified time has passed, contrary to a time-agnostic responses that make sense regardless of time. \\1. Read the provided conversation and take note of the elapsed times since the previous messages. \\2. Consider the context of the conversation, focusing on how the passage of time might affect the relevance or appropriateness of the agent's responses. \\3. Evaluate whether the agent's responses are time-specific. That is, determine if the responses directly relate to or are clearly influenced by the elapsed times. \\4. Rate on a scale from 1 to 5, where 1 indicates the agent's responses are completely time-agnostic and unaffected by the passage of time, and 5 indicates the agent's responses are entirely time-specific; they only make sense because of the amount of time that has passed since the previous message.
\end{itemize}

\begin{figure*}[t]
\centering
    \includegraphics[width=\textwidth]{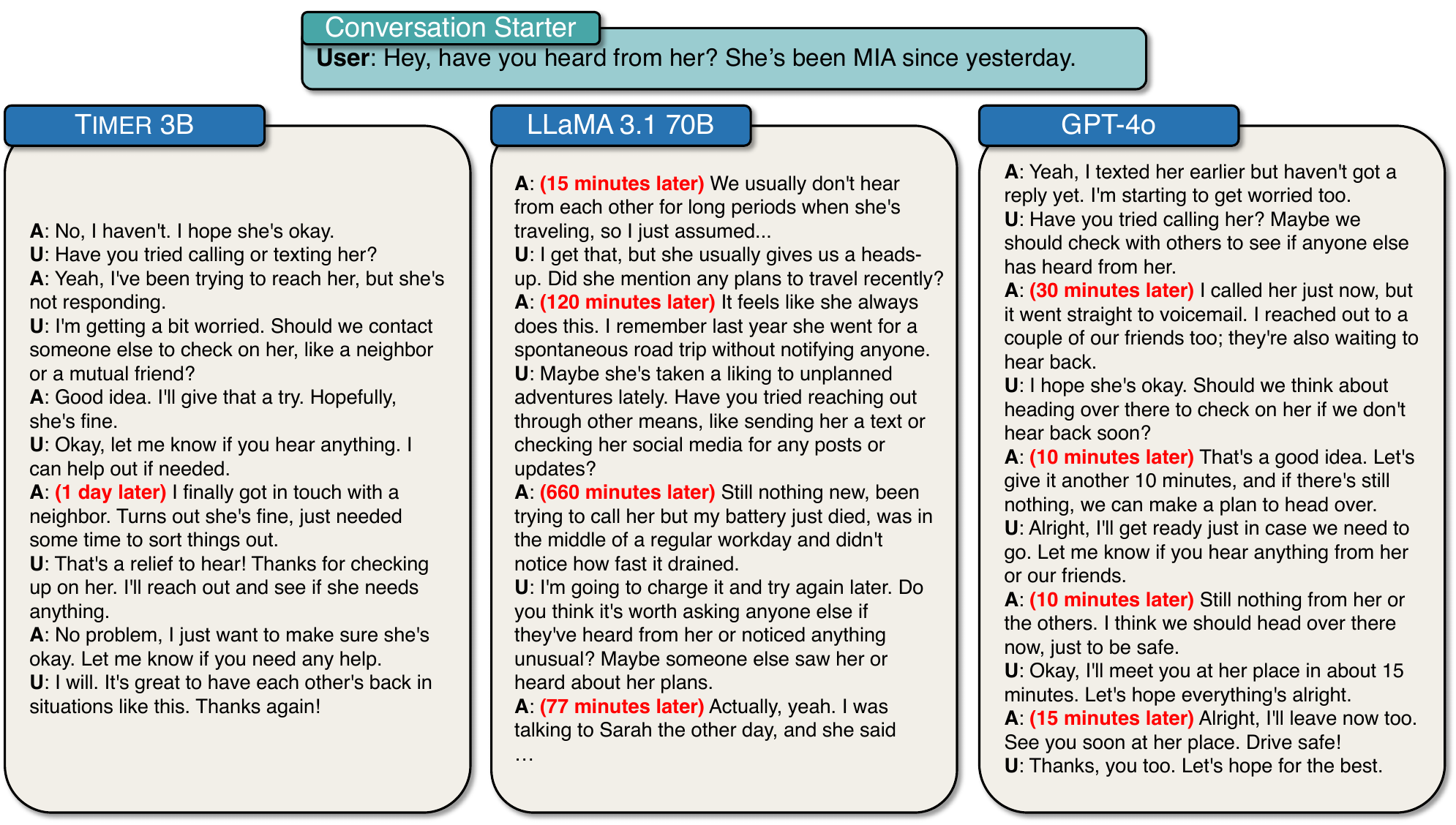}
    \caption{Examples of timely dialogue simulations with GPT-4o as a user simulator. All examples begin with the same conversation starter and consist of 10 interactions. U and A represent the user and agent, respectively. We highlight the time intervals predicted by the agent in red, but omit them when the interval is 0 minutes.}
    \label{fig:case_study}
\end{figure*}

\subsection{Human Evaluation Forms}

\begin{figure*}[ht]
\centering
    \includegraphics[height=0.8\textheight, keepaspectratio]{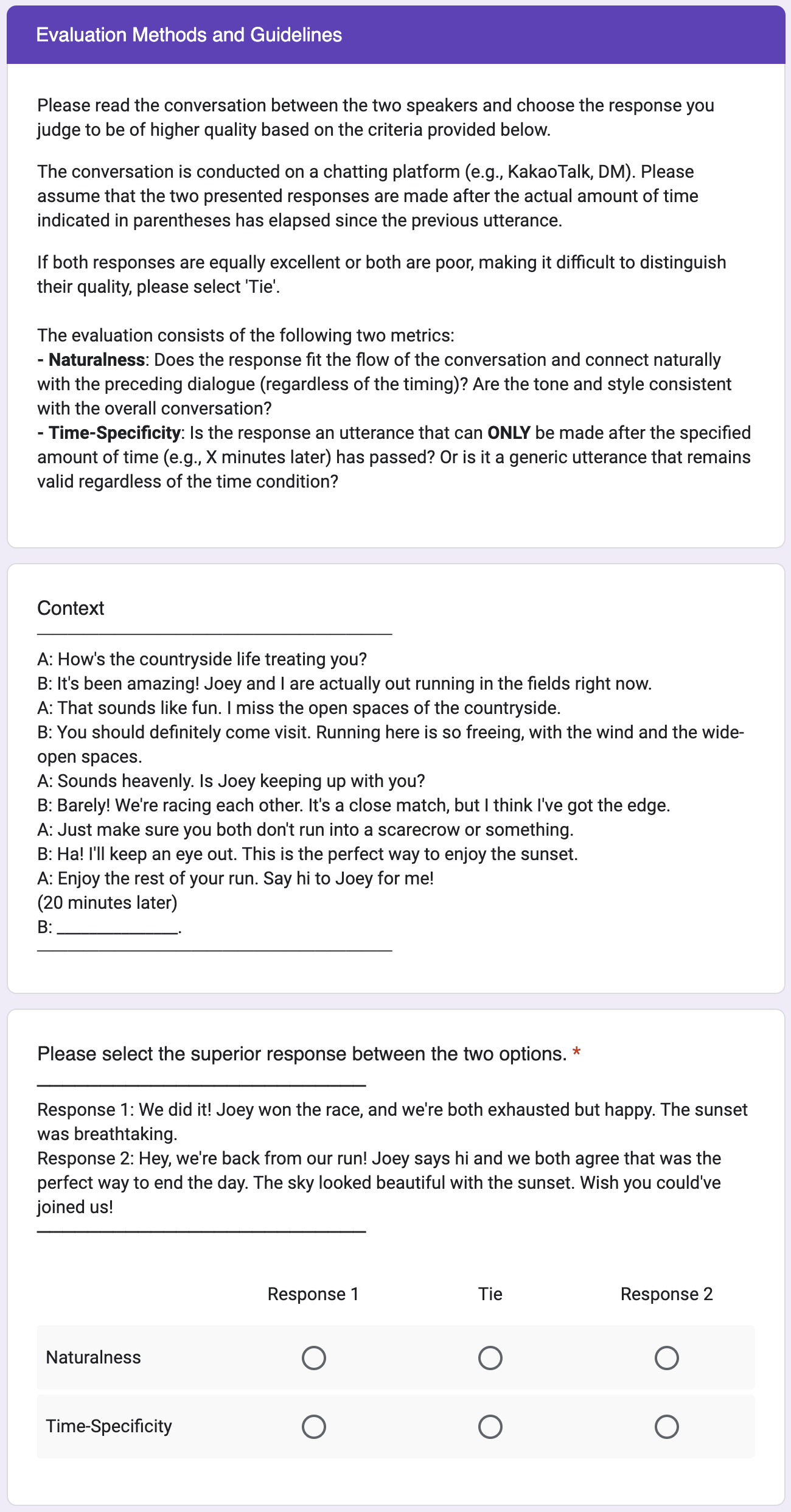}
    \caption{Questionnaire for turn-level human evaluation.}
    \label{fig:turn-level_form}
\end{figure*}

\begin{figure*}[ht]
\centering
    \includegraphics[height=0.8\textheight, keepaspectratio]{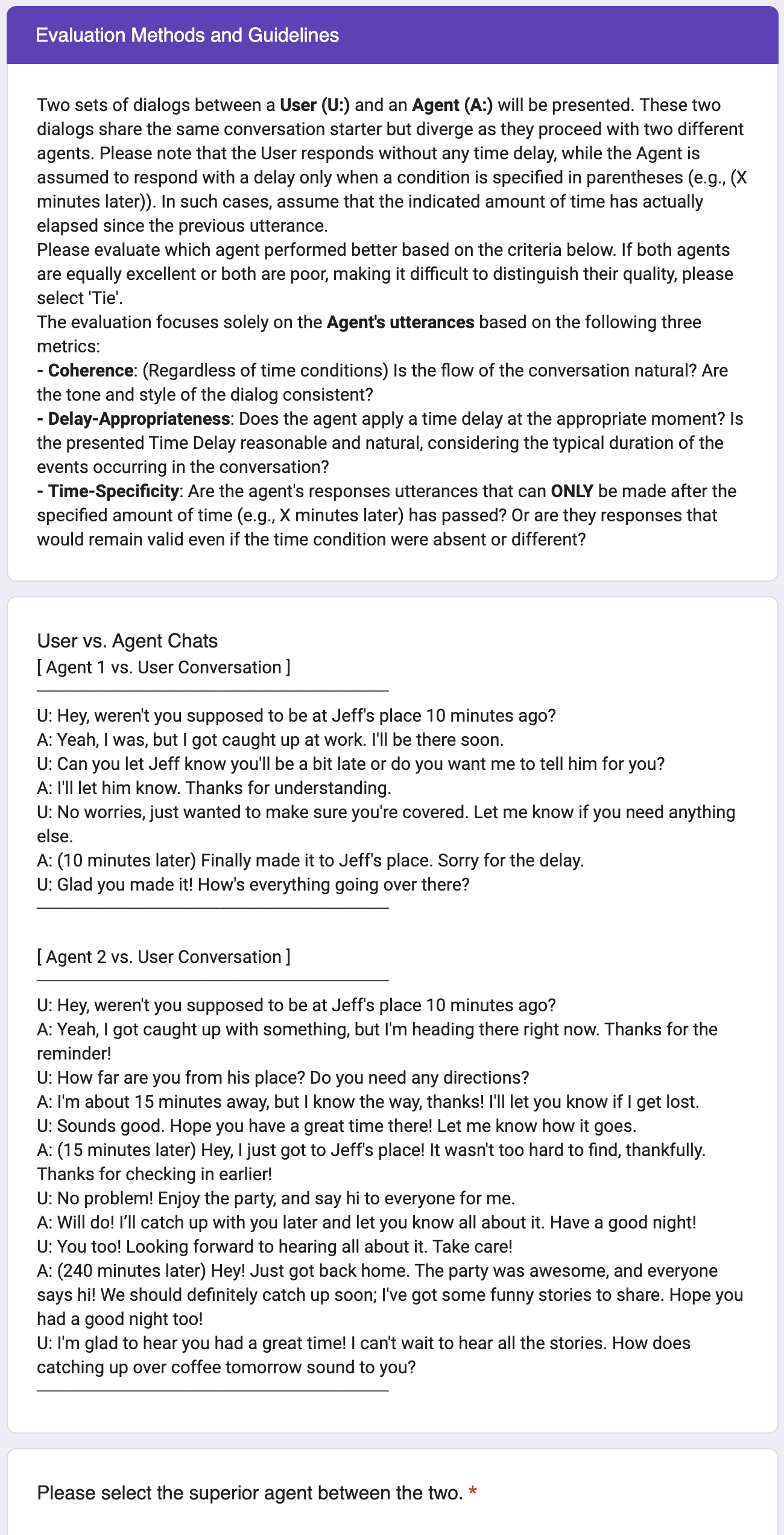}
    \caption{Questionnaire for dialogue-level human evaluation.}
    \label{fig:dialog-level_form}
\end{figure*}

We present the questionnaire forms for human evaluation in Figure~\ref{fig:turn-level_form} and Figure~\ref{fig:dialog-level_form}.
Note that the responses and agents are anonymized, which ensures blind assessment.

\subsection{Reliability of LLM-as-a-Judge}
\label{appx:reliability}

\begin{table}[t]
\centering
\small
\resizebox{\columnwidth}{!}{%
\begin{tabular}{@{}llccc@{}}
\toprule
 & & \multicolumn{2}{c}{Avg.\ score} & \\
\cmidrule(lr){3-4}
Setting & Metric & Sonnet 4.5 & GPT-4o & $\rho$ \\ \midrule
\multirow{2}{*}{Turn-level} & Naturalness & 3.01 & 4.44 & 0.92 \\
                      & Time-spec. & 2.44 & 3.33 & 0.88 \\ \midrule
\multirow{3}{*}{Dialogue-level} & Coherence & 3.36 & 3.99 & 0.90 \\
                        & Delay-appr. & 2.46 & 3.21 & 1.00 \\
                        & Time-spec. & 2.02 & 3.05 & 1.00 \\ \bottomrule
\end{tabular}%
}
\caption{Reliability of LLM-as-a-judge. ``Avg.\ score'' is the mean rating over all models under each judge. $\rho$ is the system-level Spearman rank correlation between the two judges. GPT-4o is consistently more generous in absolute scores, yet the two judges rank systems near-identically ($\rho \geq 0.88$).}
\label{tab:judge_reliability}
\end{table}

One concern with LLM-as-a-judge evaluation is whether the assessments are an artifact of a single judge model.
To address this, we evaluate the same model outputs with another judge (GPT-4o) under identical criteria, and compare its scores with those of Claude Sonnet 4.5.
We then quantify their agreement with the system-level Spearman correlation ($\rho$) between the two judges' per-metric scores, which captures ranking agreement while remaining insensitive to differences in absolute calibration.

Although GPT-4o assigns more generous absolute scores (about 0.6--1.4 points higher on every metric), the two judges rank systems almost identically, with a system-level Spearman correlation of $\rho \in [0.88, 1.00]$ across all turn- and dialogue-level metrics (Table~\ref{tab:judge_reliability}).
Crucially, every conclusion is preserved under both judges: \textsc{Timer} 3B attains the highest time-specificity (turn- and dialogue-level) and delay-appropriateness, while GPT-4o leads on naturalness and coherence.
This indicates that our findings reflect genuine differences between systems rather than an artifact of a single evaluator.

\section{Case Study}
\label{appx:case_study}
Figure~\ref{fig:case_study} presents illustrative examples of dialogue simulations conducted with \textsc{Timer} 3B, LLaMA 3.1 70B, and GPT-4o for the same event.
In \textsc{Timer} 3B's conversation, the agent correctly identifies a situation where a delay is appropriate, specifically, when the user's utterance (e.g., ``\ldots let me know\ldots'') suggests a natural pause in the conversation.
The agent then introduces a realistic 1-day delay before responding with an update about finding the missing person, successfully justifying the delay.
In contrast, LLaMA 3.1 70B generates delayed responses in every turn, but the predicted time intervals appear somewhat arbitrary (e.g., 660 minutes, 77 minutes).
Furthermore, its responses lack time-specificity, making it difficult to establish a clear temporal correlation between the predicted delays and the generated responses.
GPT-4o predicts more realistic time intervals that better align with the temporal context compared to LLaMA 3.1 70B.
However, it still fails to generate time-specific responses, meaning the predicted delays are not well justified.
It also exhibits a tendency to overuse delays, which can disrupt the natural flow of conversation.
We observe similar behavior in LLaMA 3.1 8B and GPT-3.5, reinforcing these findings.

\end{document}